\documentclass[english]{article}
\pdfoutput=1
\usepackage[preprint,nonatbib]{nips_2018_wider_nonotice}

\usepackage[T1]{fontenc}
\usepackage[latin9]{inputenc}
\usepackage{color,colortbl}
\usepackage{babel}
\usepackage{verbatim}
\usepackage{url}
\usepackage{amsmath}
\usepackage{amssymb}
\usepackage{graphicx}
\usepackage{setspace}
\usepackage{cancel}
\usepackage{hyperref}
\usepackage{cleveref}
\hypersetup{linkcolor=blue,filecolor=magenta,urlcolor=cyan} 
\usepackage{appendix}
\usepackage{courier}
\usepackage{cprotect}
\usepackage{makecell}
\usepackage{listings}
\graphicspath{{figures/}}
\graphicspath{{figures/}}

\usepackage[labelfont=bf]{caption}
\usepackage{subcaption}

\usepackage{makecell}
\usepackage{multirow}

\makeatletter
\newcommand{\be}{\begin{eqnarray}}
\newcommand{\ee}{\end{eqnarray}}

\allowdisplaybreaks \numberwithin{equation}{section}
\makeatother

\def\<{\langle}

\usepackage[disable]{todonotes} 
\usepackage{tabularx} 
\usepackage{booktabs} 

\usepackage{amsmath}
\usepackage{empheq}
\usepackage{xcolor}
\definecolor{lightgreen}{HTML}{FFFF99}

\usepackage{geometry}
\usepackage{multirow}
\usepackage{array, makecell, rotating}

\begin{document}

\title{Scaling Laws and Interpretability of Learning from Repeated Data}
\author{Danny Hernandez\thanks{Correspondence to: danny@anthropic.com \newline All authors are at Anthropic. Author contributions are listed at the end of the paper.} \\ 
\and \bf 
Tom Brown, Tom Conerly, Nova DasSarma,  Dawn Drain, Sheer El-Showk, Nelson Elhage,
\and \bf 
Zac Hatfield-Dodds, Tom Henighan, Tristan Hume, Scott Johnston,
\and \bf
Ben Mann, Chris Olah, Catherine Olsson, \\ 
\and \bf  Dario Amodei, Nicholas Joseph, Jared Kaplan, Sam McCandlish
\AND \\
{\Large Anthropic}
}

\maketitle

\begin{abstract}
Recent large language models have been trained on vast datasets, but also often on repeated data, either intentionally for the purpose of upweighting higher quality data, or unintentionally because data deduplication is not perfect and the model is exposed to repeated data at the sentence, paragraph, or document level. Some works have reported substantial negative performance effects of this repeated data. In this paper we attempt to study repeated data systematically and to understand its effects mechanistically. To do this, we train a family of models where most of the data is unique but a small fraction of it is repeated many times. We find a strong double descent phenomenon, in which repeated data can lead test loss to increase midway through training. A predictable range of repetition frequency leads to surprisingly severe degradation in performance. For instance, performance of an 800M parameter model can be degraded to that of a 2x smaller model (400M params) by repeating 0.1\% of the data 100 times, despite the other 90\% of the training tokens remaining unique. We suspect there is a range in the middle where the data can be memorized \it and\rm\space doing so consumes a large fraction of the model's capacity, and this may be where the peak of degradation occurs.  Finally, we connect these observations to recent mechanistic interpretability work --- attempting to reverse engineer the detailed computations performed by the model --- by showing that data repetition disproportionately damages copying and internal structures associated with generalization, such as induction heads, providing a possible mechanism for the shift from generalization to memorization. Taken together, these results provide a hypothesis for why repeating a relatively small fraction of data in large language models could lead to disproportionately large harms to performance.

\end{abstract}

\setcounter{footnote}{0} 

\section{Introduction}

\begin{figure}
    \centering
    \includegraphics[width=\linewidth]{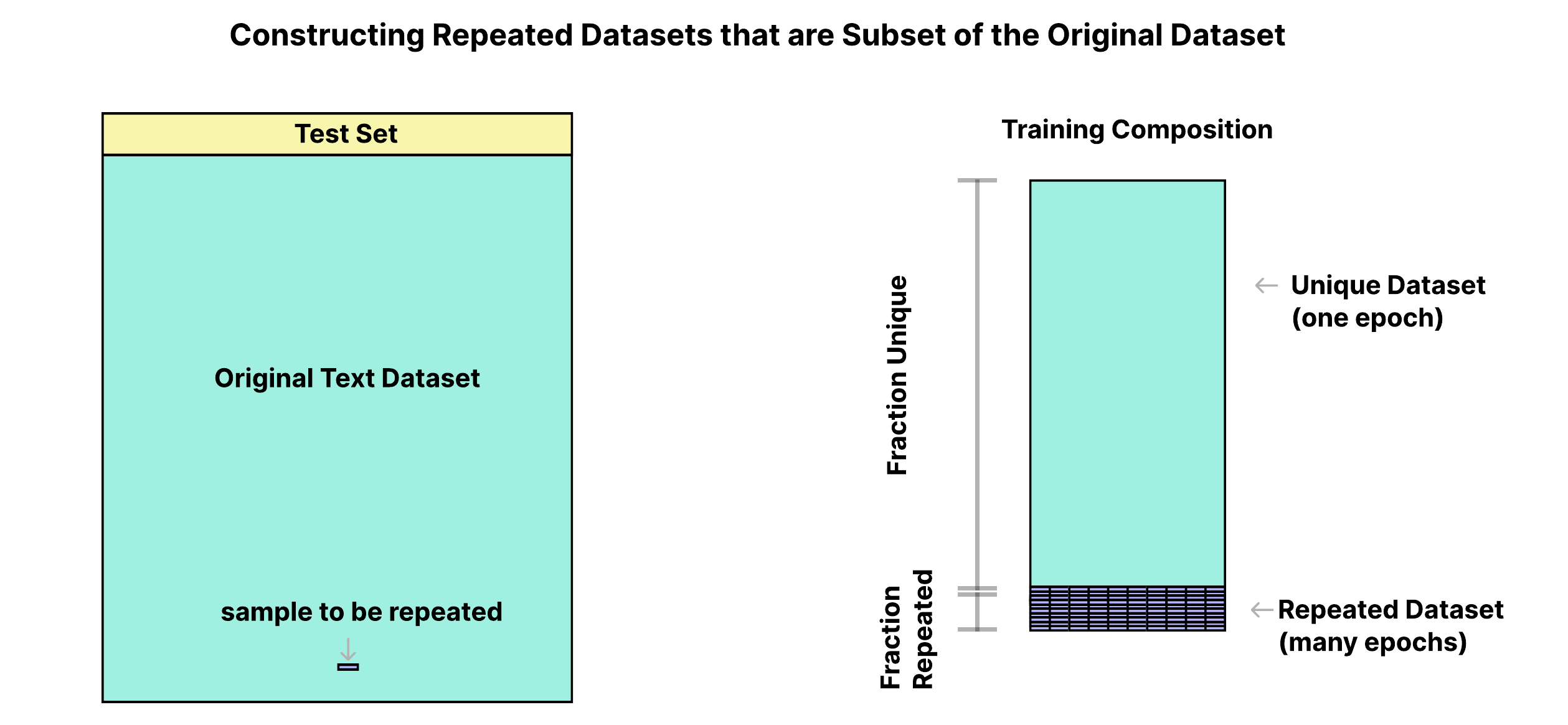}
    \caption{Experimental Setup. From a large original text dataset (left), we draw 90\% of our desired training dataset in a non-repeated fashion, and 10\% as repeats of a tiny portion of the original dataset (right). We hold constant that 10\% of total training tokens will come from repeats, but we vary the repeated fraction in our runs. In other words, the sample to be repeated might be very small, like 0.01\% of the total training tokens repeated 1000x, or relatively large, like 1\% of the total training tokens repeated 10x. A small, held-back portion of the original dataset (yellow in left figure), not including any repeated data, is used as a test set and is the test loss reported in all subsequent figures.
    }
    \label{fig:setup}
\end{figure}
Large, high-quality text datasets are crucial for training large language models \cite{https://doi.org/10.48550/arxiv.2005.14165, https://doi.org/10.48550/arxiv.2112.11446}. Such datasets often contain many copies of substantially overlapping documents, which greatly impairs the performance of language models on downstream tasks \cite{lee2021deduplicating}. However, it is not well understood why data repetition impacts performance to such a large extent.

In this paper we study data repetition in language models through two lenses: the macroscopic lens of scaling laws, and the microscopic lens of mechanistic interpretability \cite{elhage2021mathematical, olsson2022context}. For the first lens, we trained transformer \cite{https://doi.org/10.48550/arxiv.1706.03762} language models on mostly unique data plus a small fraction of repeated data (Figure \ref{fig:setup}), varying the repeated dataset size, model size, and fraction of tokens trained on repeated data. We find a strong double-descent phenomenon \cite{https://doi.org/10.48550/arxiv.1710.03667, https://doi.org/10.48550/arxiv.1812.11118, https://doi.org/10.48550/arxiv.1912.02292}, such that there is a defined range of repetition frequency for which performance is harmed to a surprisingly large extent. We suspect there is a range in the middle where the data can be memorized \it and\rm\space doing so consumes a large fraction of the model's capacity, and this may be where the peak of degradation occurs. The location of the region suggests that large models like GPT-3, Gopher, and PALM \cite{https://doi.org/10.48550/arxiv.2005.14165, https://doi.org/10.48550/arxiv.2112.11446, https://doi.org/10.48550/arxiv.2004.07159} need to be careful about overfitting their high quality distributions like Wikipedia and books.

For the second lens, mechanistic interpretability (attempting to reverse engineer the detailed computations performed by the model) we show that repeated data disproportionately damages induction heads. Induction heads use a circuit of 2 attention heads to "complete the pattern by copying and completing sequences" \cite{olsson2022context}. The damage to induction heads is observed through degradation in copying, prefix matching, and through inspection.

Together, the two lenses provide an integrated picture of how repeated data might be causing the network (or part of it) to shift from generalization to memorization, and mechanistically how this could be harming performance of the overall language model.

\subsection{Summary of Results}

\begin{figure}
    \centering
    \includegraphics[width=0.49\textwidth]{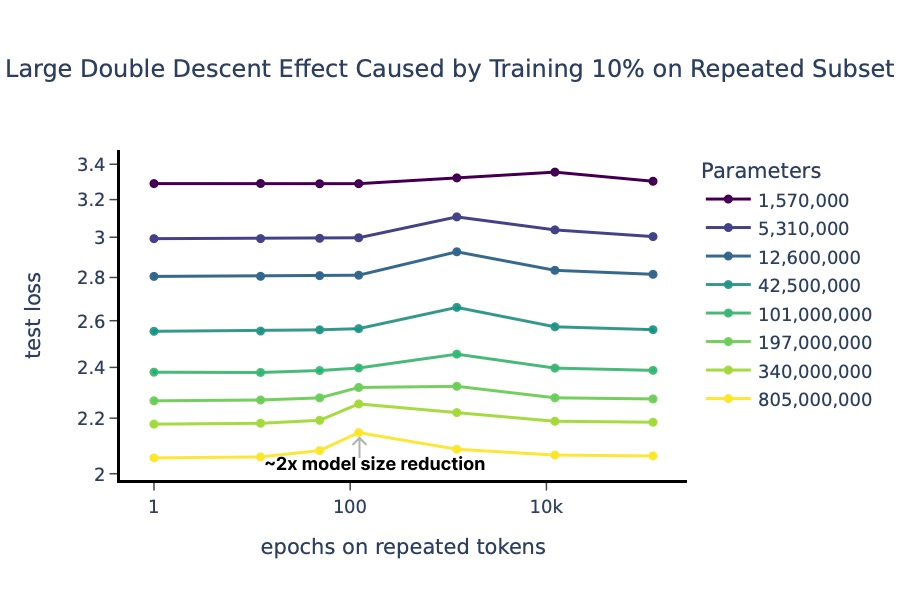}
    \includegraphics[width=0.49\textwidth]{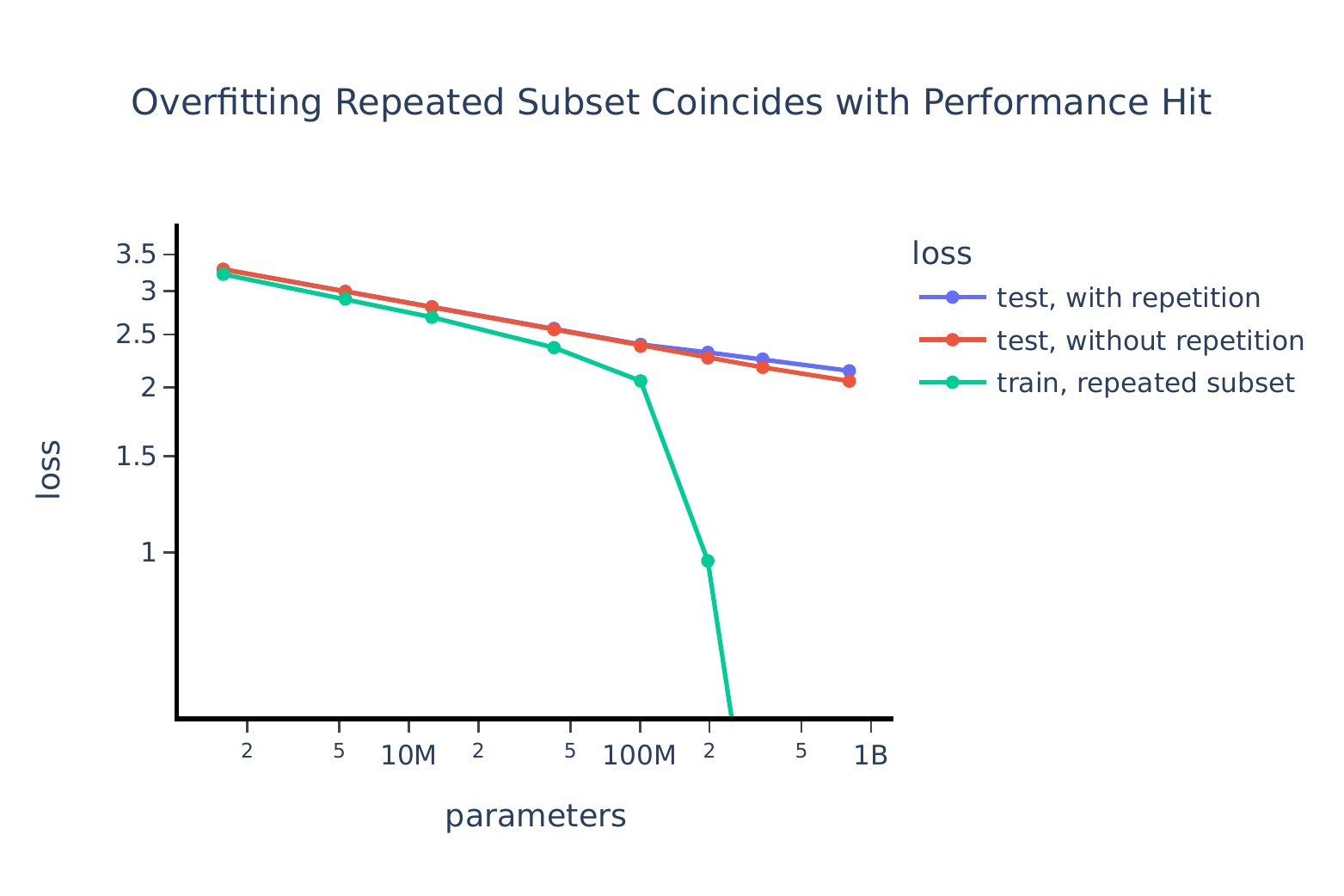}
    \caption{Models of different sizes show a degradation in performance at a specific range of repeats that shrinks with model size (left panel). At its peak the degradation sometimes reaches the equivalent of a 2x decrease in model size. The right panel shows that divergence (blue line) from a healthy, straight scaling law (red) lines up with when the models start to dramatically overfit the repeated subset (green curve). The blue line on the right corresponds to a vertical slice of models in the left diagram trained on the repeated subset for 120 epochs. All these models were trained on 90\% unique data and 10\% repeated tokens.}
    \label{fig:double_descent}
\end{figure}

To systematically study repeated data, we trained transformer \cite{https://doi.org/10.48550/arxiv.1706.03762} language models on mostly unique data plus a small fraction of repeated data (Figure \ref{fig:setup}), varying the repeated dataset size, model size, and fraction of tokens trained on repeated data over 2-3 orders of magnitude. All models were trained for 100B tokens. We examined the resulting models using both scaling laws and mechanistic interpretability tools. Our main findings were as follows:

\begin{itemize}
\item \textbf{Repeated data induces a strong double-descent phenomenon} \cite{https://doi.org/10.48550/arxiv.1710.03667, https://doi.org/10.48550/arxiv.1812.11118, https://doi.org/10.48550/arxiv.1912.02292}, in which data repeated a few times does not cause much damage to language model performance, data repeated very many times also does not cause much damage, but there is a peak in the middle where damage is surprisingly large. For instance, when we train an 800M parameter transformer with 10\% of training tokens drawn from the repeated subset (yellow curve in Figure \ref{fig:double_descent}) we find the loss can be nearly as high as for the 340M parameter transformer (light green curve). We see an epoch-wise \cite{https://doi.org/10.48550/arxiv.1912.02292} double descent learning curve in Figure \ref{fig:dd_learning_curve} is driving this performance degradation. We suspect there is a range in the middle where the data can be memorized \it and\rm\space doing so consumes a large fraction of the model's capacity, and this may be where the peak of degradation occurs. Figure \ref{fig:double_descent} on the right shows that the peak performance hit coincides with where the train loss on the repeated data approaches zero, similar to previously observed double-descent phenomena. This also provides a practical diagnostic for when repeated data is likely to be harming the model.
\item \textbf{Repeated data can cause a divergence from power-law scaling.} For the blue curve in Figure \ref{fig:double_descent} right (122 repeated epochs), we see only a moderate impact to performance (line on log-log graph) until the model is scaled up to 100M parameters, after which we see a large divergence from power law scaling of cross entropy loss. Extrapolating the region of large degradation in Figure \ref{fig:poor_scaling} predicts meaningful degradation of repeating data only 2 times for large (GPT-3 size) models, though the region would be shifted if the models were trained to the compute optimal frontier \cite{https://doi.org/10.48550/arxiv.2203.15556}.
\item \textbf{Repeated data causes a disproportionately large performance hit to copying, a mechanism for in-context learning.} We constructed a simple copying eval, the loss on the first paragraph of Harry Potter copied 11 times. We observe that using 3\% repeated data at the worst number of repeated epochs caused up to a 3x reduction in effective model size (performance equal to model with 3x fewer parameters) on this task whereas it only caused at most a 15\% reduction in effective model size on test loss.
\item \textbf{The disproportionate performance hit to copying coincides with a disproportionate degradation of induction heads}. In line with \cite{olsson2022context} we evaluated the models on their prefix matching score, repeated sequences of random tokens and observed the degree to which attention heads attend to earlier tokens that are preceded by a token that matches the present token. We observe that using 3\% repeated data at the worst number of repeated epochs caused on average a 32\% reduction in effective model size on this task whereas it only caused at most a 15\% reduction in effective model size on test loss.
\item \textbf{Repeated text data causes a small but still disproportionate performance drop out of distribution, as measured by cross entropy loss on Python code.} Unlike our the Harry Potter copying and prefix matching evals we mostly see the performance drop with higher levels of repetition, 50-90\%.
\item \textbf{One and two-layer attention only models trained on repeated data are worse at exactly copying and fuzzily copying (for instance correctly predicting Dursleys given that Dursley has appeared previously) proper names on inspection.} When we inspect per tokens losses of smaller models we can see this degradation in a simple, understandable form of copying in a paragraph of text.
\item \textbf{Training on repeated Python code creates a similar behavior.} When training on Python we also observe a double descent phenomenon and a predictable poor performance region in terms of model size and repeated epochs, though the shape of both curves are somewhat different.
\item \textbf{Pre-training on repeated data damages models.} Pre-training with repeated data leads to worse performance than both training from scratch and fine-tuning from a control model pre-trained on the original text dataset. During fine-tuning, the repeated data model forgets the repeated dataset, so we consider the model pre-trained with repeated data to be strictly worse than the model fine-tuned from the unique dataset.
\end{itemize}

\section{Results}


\begin{figure}
    \centering
    \includegraphics[width=0.49\textwidth]{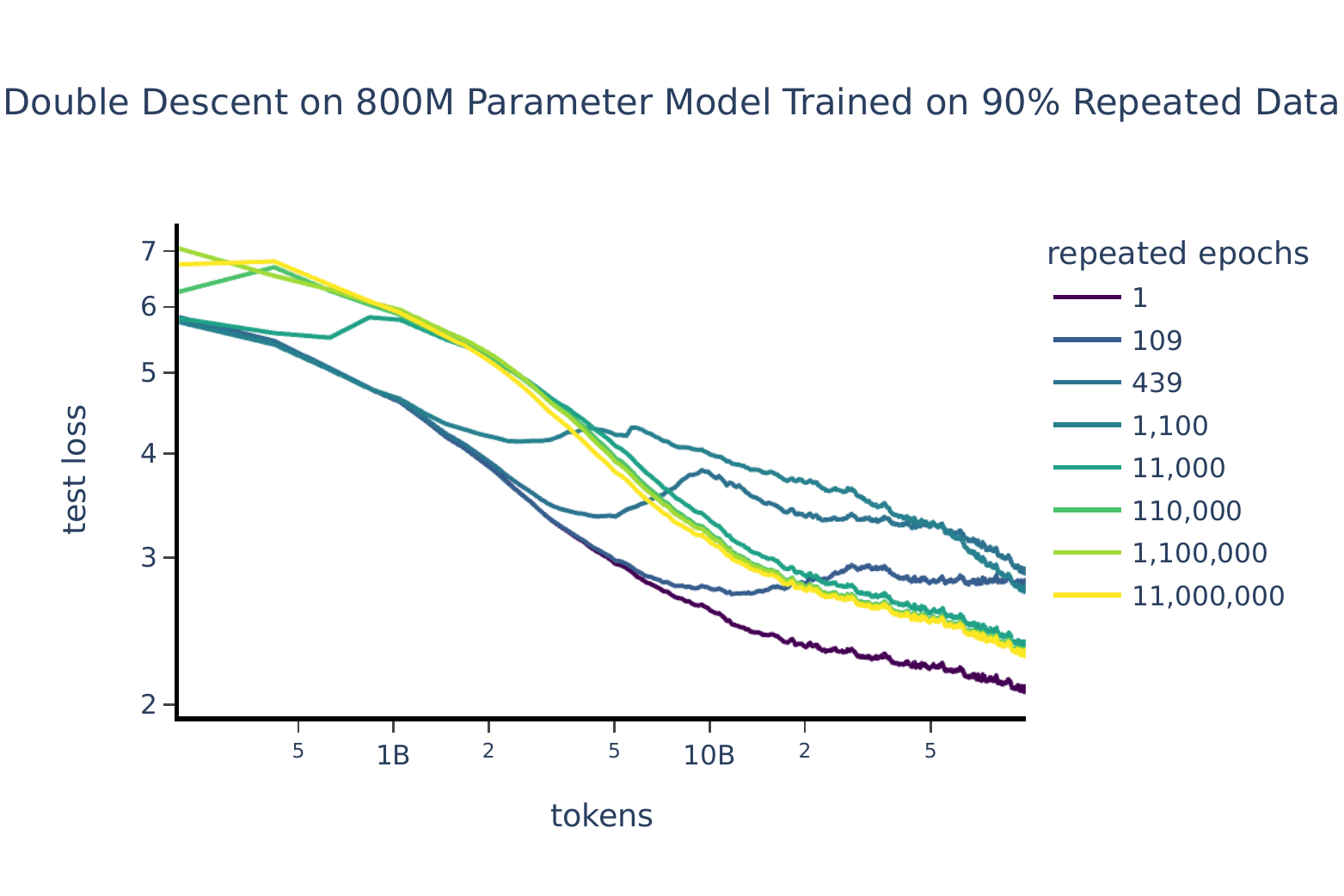}
    \includegraphics[width=0.49\textwidth]{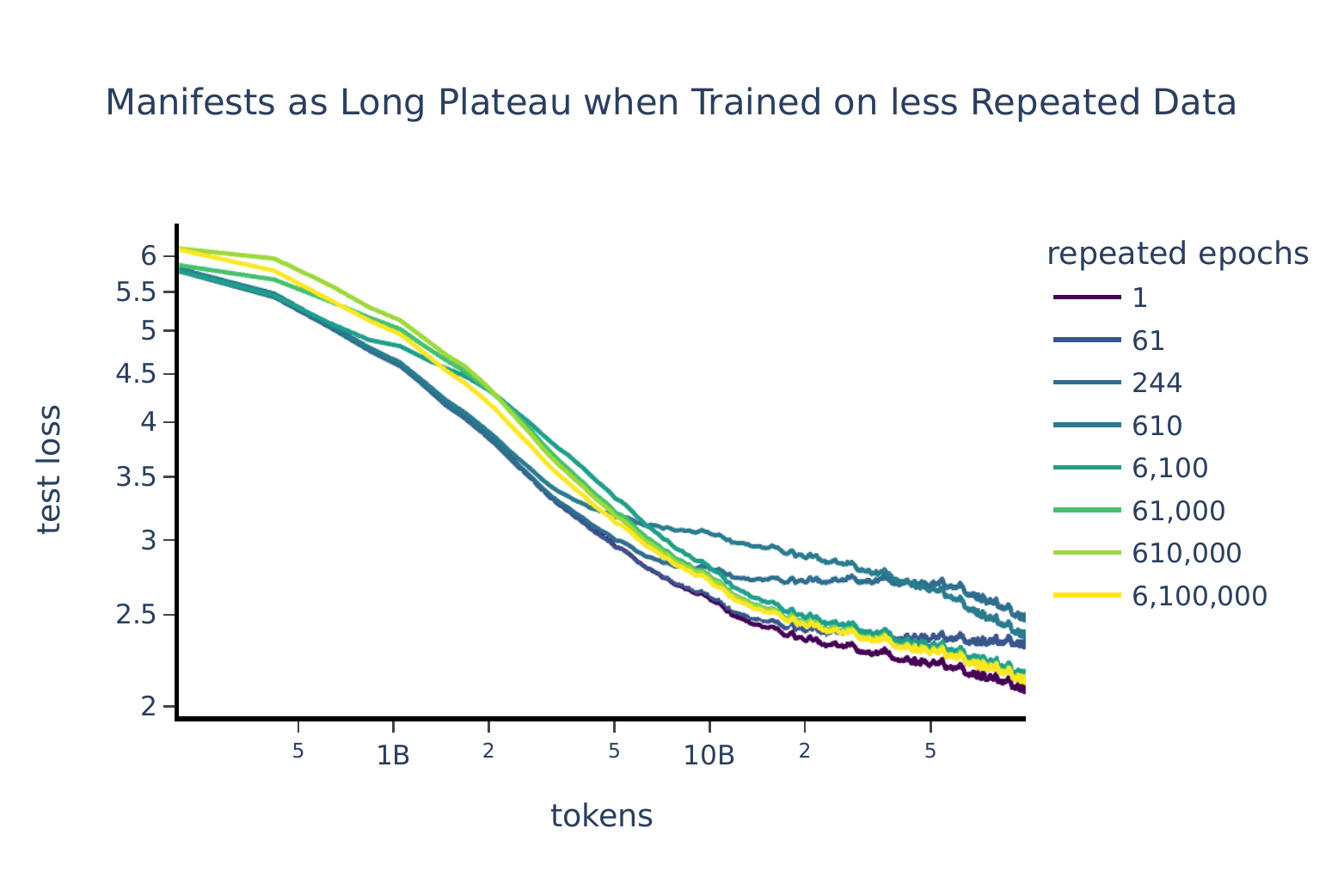}
    \caption{ Learning curves for test loss on 800M models with 90\% repeated data (left) and 50\% repeated data (right), each with varying numbers of repeats/sizes of the repeated fraction. The graph on the left shows characteristic double descent curves. Repeated epochs corresponds to the number of epochs on the repeated tokens, the rest of the data is seen only once. For several models, test loss drops as normal during the beginning of training, but then starts to rise during the middle of training before dropping again. In the graph on the right with only 50\% repeated data, we see that the double descent bumps have turned into long plateaus for highly affected models.}
    \label{fig:dd_learning_curve}
\end{figure}

\textbf{Repeated data induces a strong double descent phenomenon.} 
The results from training models on different sizes, fractions of repeated data, and frequency of repeats are shown in Figures 2 and 3.  Figure 2 (left) shows that when we train on 10\% repeated data and vary the frequency of repetition (or equivalently the number of epochs of repeated data), there is a specific range of repetition frequency for which damage to model performance is maximized.  The range depends on the model size but for a 800M parameter model it occurs at roughly 100x repeats of 0.1\% of the data, and degrades performance nearly to that of a 340M parameter model.  This is a large degradation given that only 10\% of the data is repeated. The peak coincides with the advent of memorization on the repeated data (Figure 2 right) -- a possible indicator of a double descent phenomenon.

Figure \ref{fig:dd_learning_curve} shows learning curves for different repetition frequencies and for 50\% and 90\% of the data being repeated.  In the extreme case of 90\% repeated data and the correct frequency of repetition (100x-10,000x), we confirm the presence of a literal double descent curve in which the loss decreases, increases, and then decreases again (Figure \ref{fig:dd_learning_curve} left).  As we lower the fraction of repeated data to 50\%, the curve becomes a long plateau rather than double descent, but it appears to be fundamentally an epoch-wise double descent phenomenon \cite{https://doi.org/10.48550/arxiv.1912.02292}. These peaks and plateaus again coincide with the training loss on the repeated data approaching zero as shown in Figure \ref{fig:double_descent}. As in \cite{https://doi.org/10.48550/arxiv.1912.02292} we see double descent effects caused by both increasing model size and epochs. We suspect there is a range in the middle where the data can be memorized \it and\rm\space doing so consumes a large fraction of the model's capacity, and this may be where the peak of degradation occurs, for a more thorough discussion of this question see the discussion (section \ref{discussion}).

\begin{figure}
    \centering
    \includegraphics[width=0.49\textwidth]{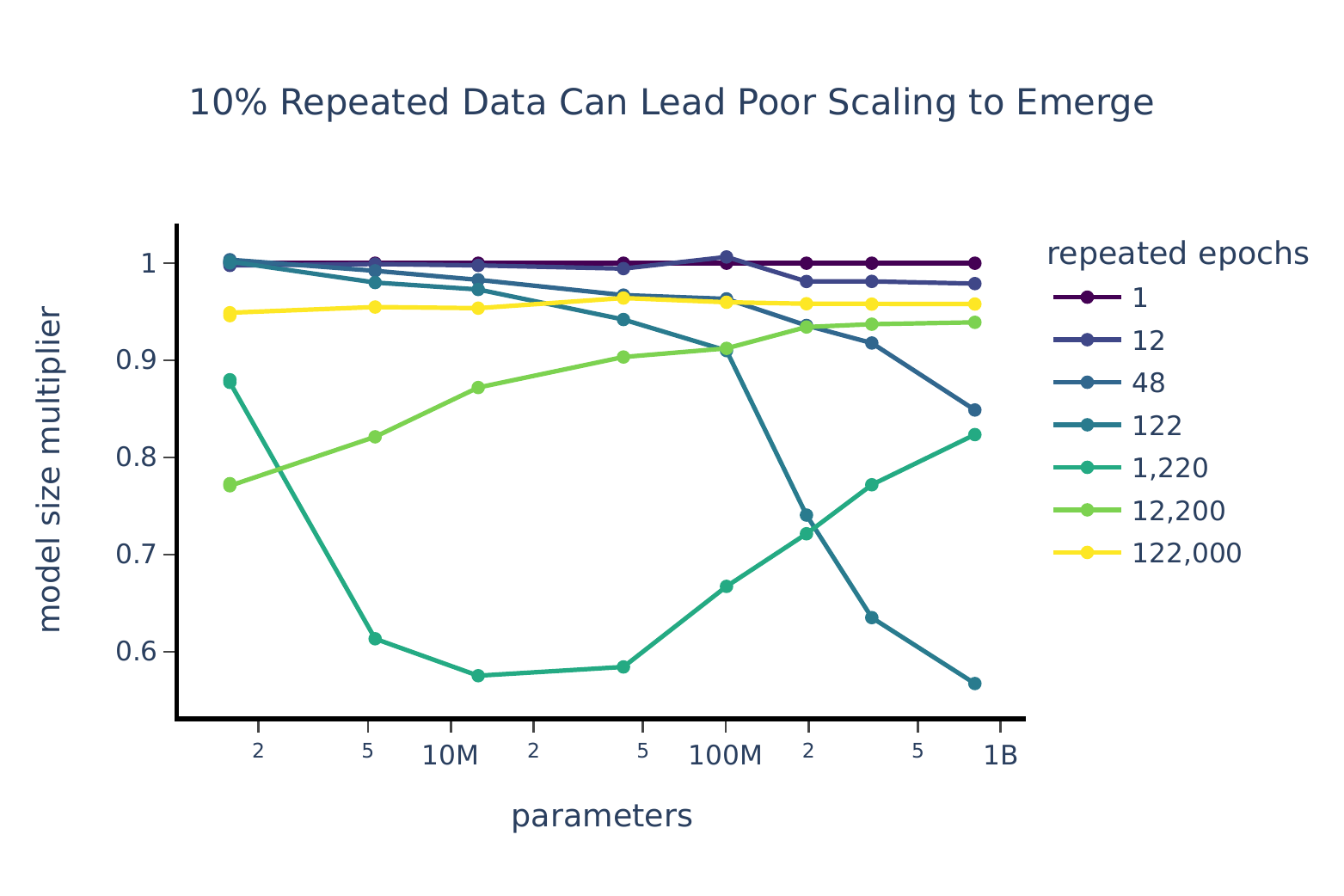}
    \includegraphics[width=0.49\textwidth]{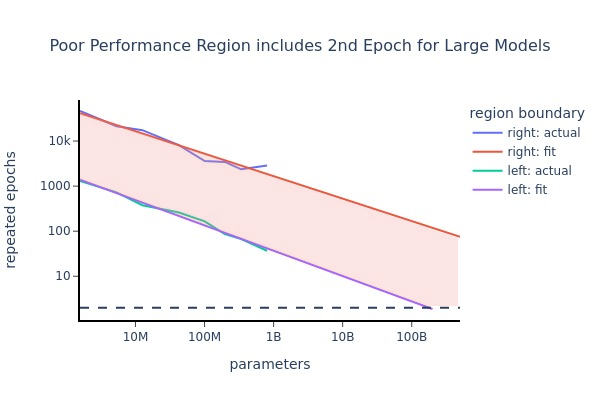}
    \caption{On the left we plot the same results as in Figure \ref{fig:double_descent}, re-parameterized in terms of the effective model size multiplier implied by the test loss (performance equal to a model with x times as many parameters). For a given number of repetitions, degradation occurs only for a specific range of model sizes. For example, for the blue curve (122 repeated epochs), we see almost no performance deviation from a power law scaling law (line on log-log graph) until the model is scaled up to 100M parameters, after which we see a divergence. We see the same divergence around 400M parameters for 12,200 repeated epochs. The right graph shows a large, predictable region over which the degradation occurs, and suggests that large models like GPT-3, Gopher, and PALM \cite{https://doi.org/10.48550/arxiv.2005.14165, https://doi.org/10.48550/arxiv.2112.11446, https://doi.org/10.48550/arxiv.2004.07159} need to be careful about overfitting their high quality distributions like Wikipedia and books -- although note that this holds constant the number of total training tokens. The blue and green curves correspond to the right and left sides of the double descent region where we observe 50\% of the maximum effect. They are an aggregation of that curve for the scans where we trained on 3\%, 10\%, 20\%, 50\%, and 90\% repeated data. The details of both fits are in Appendix \ref{app:region_fits}. A large number of runs needed to be aggregated to produce a clean fit for region of reduced performance.}
    \label{fig:poor_scaling}
\end{figure}

\textbf{Repeated data can cause a divergence from power-law scaling.} Figure \ref{fig:poor_scaling} zooms in on the degradation of performance, measured as a function of model size for different repetition frequencies of the repeated data.  For example, models trained for 1,220 repeats and 10\% repeated data show a dip in performance to the equivalent of a model 0.55x as large, when the model size is 10M to 100M parameters.  As the model size continues to increase, performance recovers to 0.8x model-size equivalent for a 1B parameter model.  For a smaller number of repeats (122 repeats), the dip occurs later, centered around 1B parameters.

The right panel of Figure \ref{fig:poor_scaling} shows the range over which we observe at least 50\% of the maximum degradation; this corresponds to a ``band'' or region in the (model size, repetition frequency) plane.  Both boundaries of the region are a good fit to a power law relating frequency of repetition to the number of parameters of the model, namely:

\[E = k * N ^ \alpha \]

where $E$ corresponds to epochs of repetition and $N$ corresponds to the parameters in the model. it is notable that the lines in figure 2b are relatively parallel. The fits for the above lines are given in the table below:

\begin{center}
\begin{tabular}{|c | c | c |} 
 \hline
  & $k$ & $\alpha$ \\
 \hline
 right boundary & 5.1e7 & -.50 \\ 

 \hline
 left boundary & 4.2e6 & -.56 \\ [1ex] 
 \hline
\end{tabular}
\end{center}

Note that extrapolating these boundaries leads to a prediction of significant degradation from repeating data as little as 2x on state-of-the-art language models with hundreds of billions of parameters, although this applies for a constant number of training tokens (100B). In practice large models are trained for more than this\cite{https://doi.org/10.48550/arxiv.2203.15556}, and as shown in Figure \ref{fig:dd_learning_curve}, training past the double descent peak is helpful, so the degradation would likely not be quite as bad. When looking at Figure \ref{fig:dd_learning_curve} we see that the the poor performance region would be shifted left for large models trained on the compute efficient frontier (the pareto frontier of compute and performance) \cite{https://doi.org/10.48550/arxiv.2001.08361}.

Overall it seems that in addition to being robust to task, model size, and architecture as shown in previous work \cite{https://doi.org/10.48550/arxiv.1710.03667, https://doi.org/10.48550/arxiv.1812.11118, https://doi.org/10.48550/arxiv.1912.02292} double descent as a general phenomenon appears to be robust to occurring in a sub-distribution and that it can have a large effect on overall performance even while being a modest fraction of training tokens.

\textbf{Repeated data causes a disproportionately large performance hit to copying, a mechanism for in-context learning}. The ability of a language model to copy text (in the sense of being provided with a context consisting of a passage repeated several times, and testing whether the model can repeat it once more) is a potential measure of \textit{generalization}, as copying is independent of the content of the text.  Also, recent interpretability work has suggested that copying may be implemented by crisp internal algorithmic structures (\cite{olsson2022context}), again suggesting generalization.  It thus seems valuable to investigate what happens to copying during a memorization-related degradation in performance, which we have shown above occurs in our experiments.

\begin{figure}
    \centering
    \includegraphics[width=0.49\textwidth]{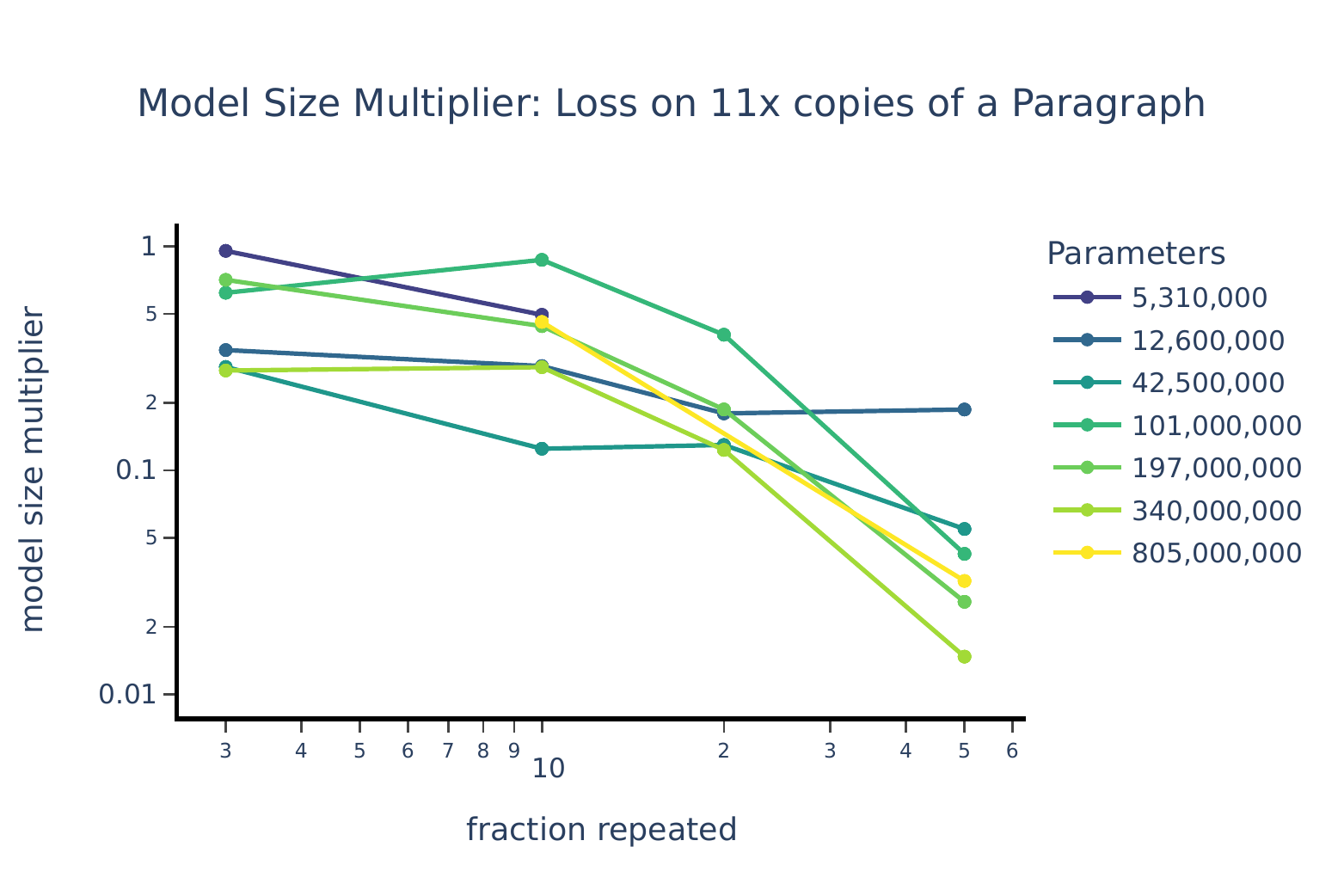}
    \includegraphics[width=0.49\textwidth]{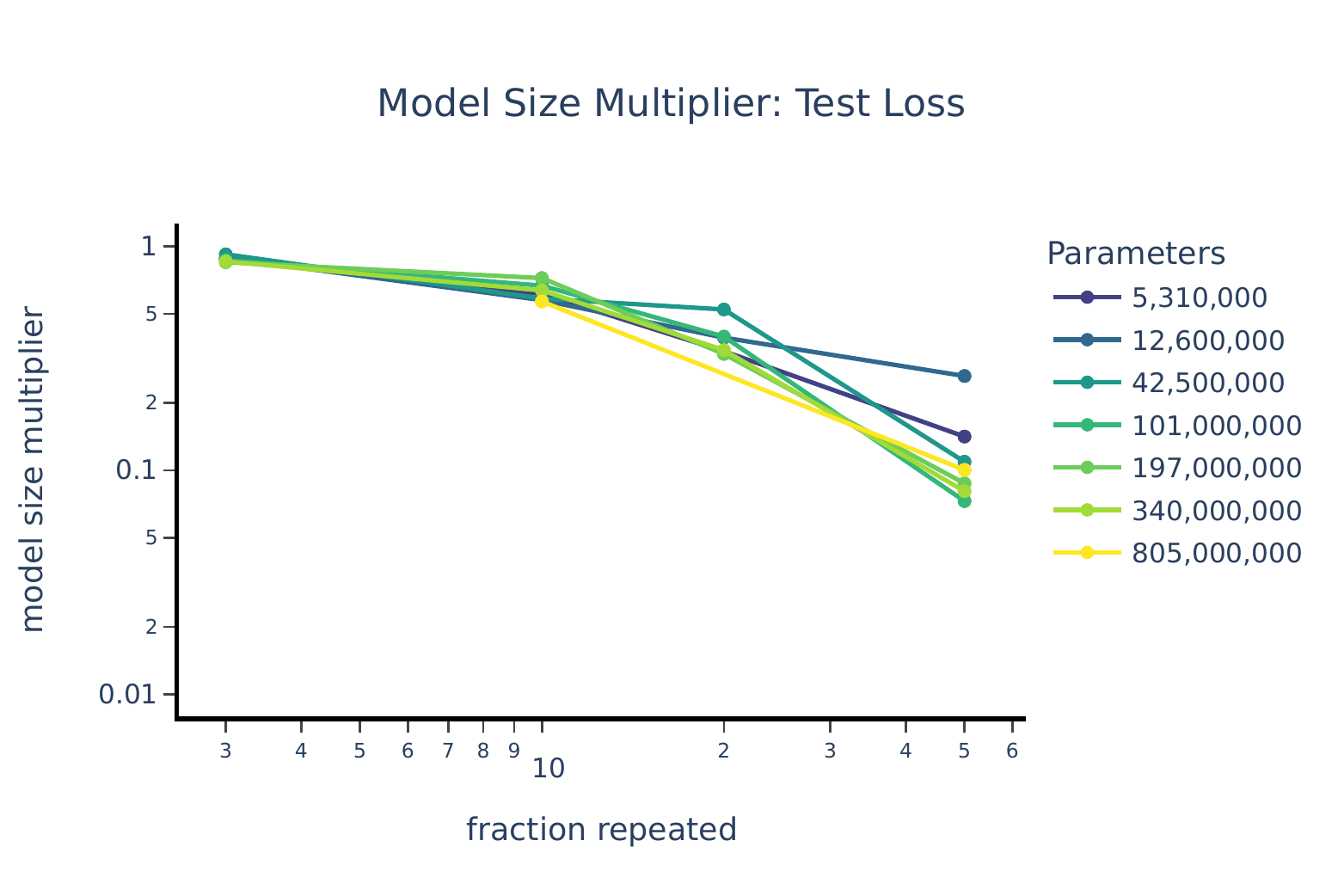}
    \caption{We constructed a simple measure of the model's copying ability, consisting of the loss on the first paragraph of Harry Potter repeated 11 times. We measured the double descent peak performance for a given model size and fraction of repeated data and compared that to a fit of these evaluations on the control model (trained on unique text) scan to generate an effective model size. We observe that 3\% repeated data at the pessimal number of repeated epochs caused a 3x reduction in effective model size on this task for a for several model sizes, whereas it only caused at most a 1.15x reduction in effective model size on test loss. We see much larger effects on the copying evaluation than on overall performance for repeated data fractions between 3\% and 20\%. The model size multiplier for copying is based on interpolation and the model size multiplier for test loss is based on a power law fit (see Appendix \ref{app:copy_prefix} for more details). }
    \label{fig:hp_copying_1}
\end{figure}

To do this constructed a simple evaluation in which copying is heavily emphasized: we measure the loss on the first paragraph of Harry Potter copied 11 times. The models trained on repeated data performed much worse on this evaluation (Figure \ref{fig:hp_copying_1}), \textit{substantially out of proportion to the degradation on the loss itself}.  In other words, copying is \textit{preferentially} harmed by training on repeated data.  For example, a 3\% fraction of repeated data leads to a 1.15x reduction in effective model size (performance equal to model with 1.15 fewer parameters) on the general loss, but a much larger 3x effective model size reduction in terms of copying ability.  As can be seen in Figure 5, the damage to copying is greater than the damage to overall loss across the entire range of repeated data fractions.  This suggests that the shift to memorization caused by repeated data is selectively harming at some behaviors associated with generalization.

To get another view on the same phenomenon, we measured the loss of various models on the $X$th consecutive copy of the Harry Potter paragraph, where $X$ runs from 1 to 12. As shown in Figure \ref{fig:hp_copying_2} (left), for most models the loss gradually decreases with increasing numbers of copies of the paragraph (i.e. the model has an easier time predicting an additional copies after seeing more consecutive copies), but at the peak of the double descent phenomenon, the loss is much higher and, strikingly, does not decrease at all with additional copies of the paragraph.  This large aberration shows how strong the selective effect of the double descent phenomenon on copying is.  General in-context learning is also harmed at the pessimal number of repeated epochs (Figure \ref{fig:hp_copying_2} right), though to a lesser extent than copying.

\textbf{The disproportionate performance hit to copying coincides with a disproportionate degradation of induction heads}.
Having connected the damage associated with repeated data with a measure of generalization (in-context copying of text), we next took the connection one step further, by trying to also probe the potential mechanistic basis of copying.  \cite{olsson2022context} identifies ``induction heads'' as a possible basis for copying and in-context learning behavior in general, so we decided to measure these and try to connect them back to the repeated data double descent phenomenon.

\cite{olsson2022context} defines induction heads by their ability to facilitate simple copying given a repeated random sequence of tokens (though in practice this definition ends up including heads with more complex behaviors too). Induction heads use a circuit of 2 attention heads to "complete the pattern by copying and completing sequences." This can be split up into attending to the relevant token (prefix matching) and increasing the logit corresponding to the attended-to token.

\begin{figure}
    \centering
    \includegraphics[width=0.49\textwidth]{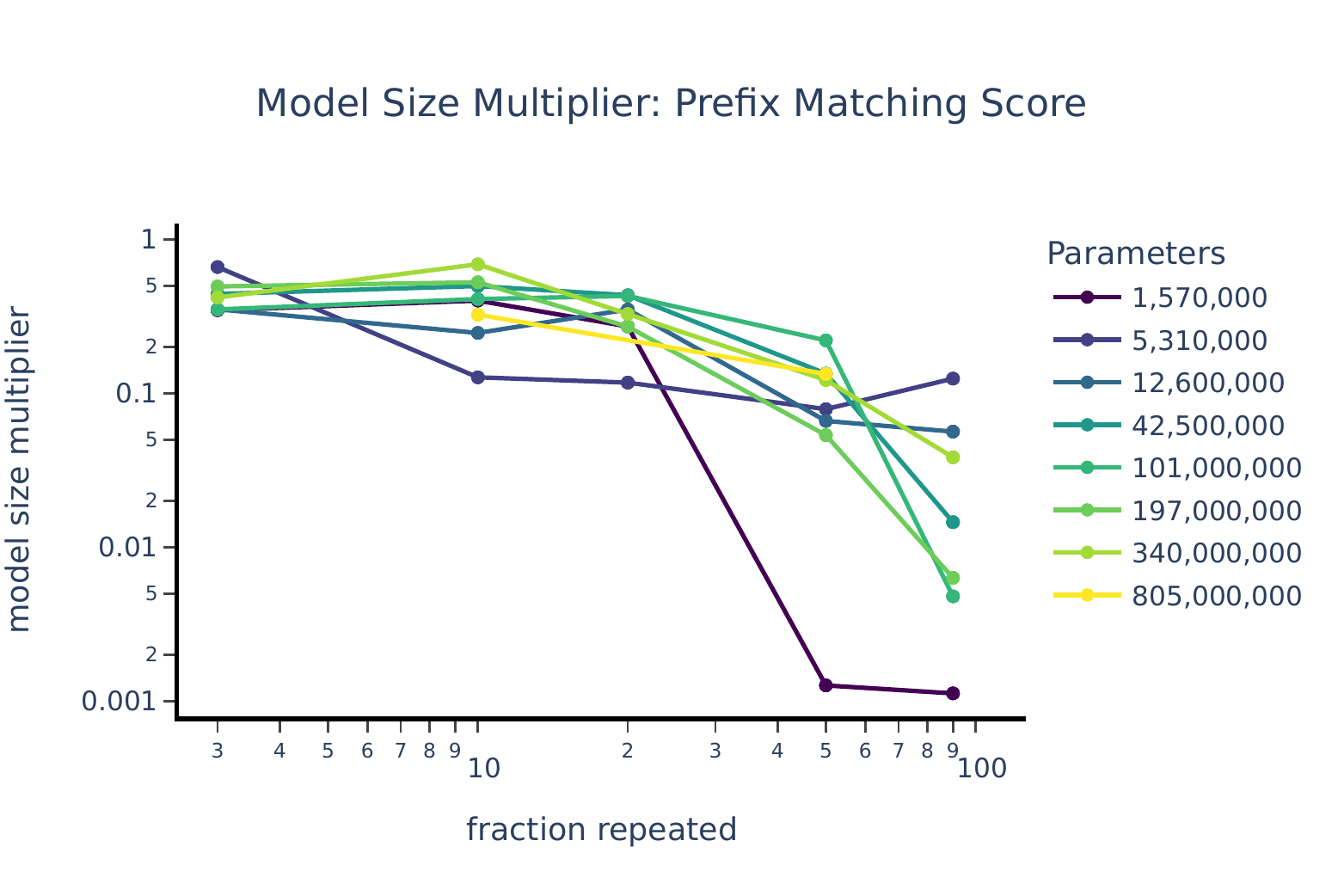}
    \includegraphics[width=0.49\textwidth]{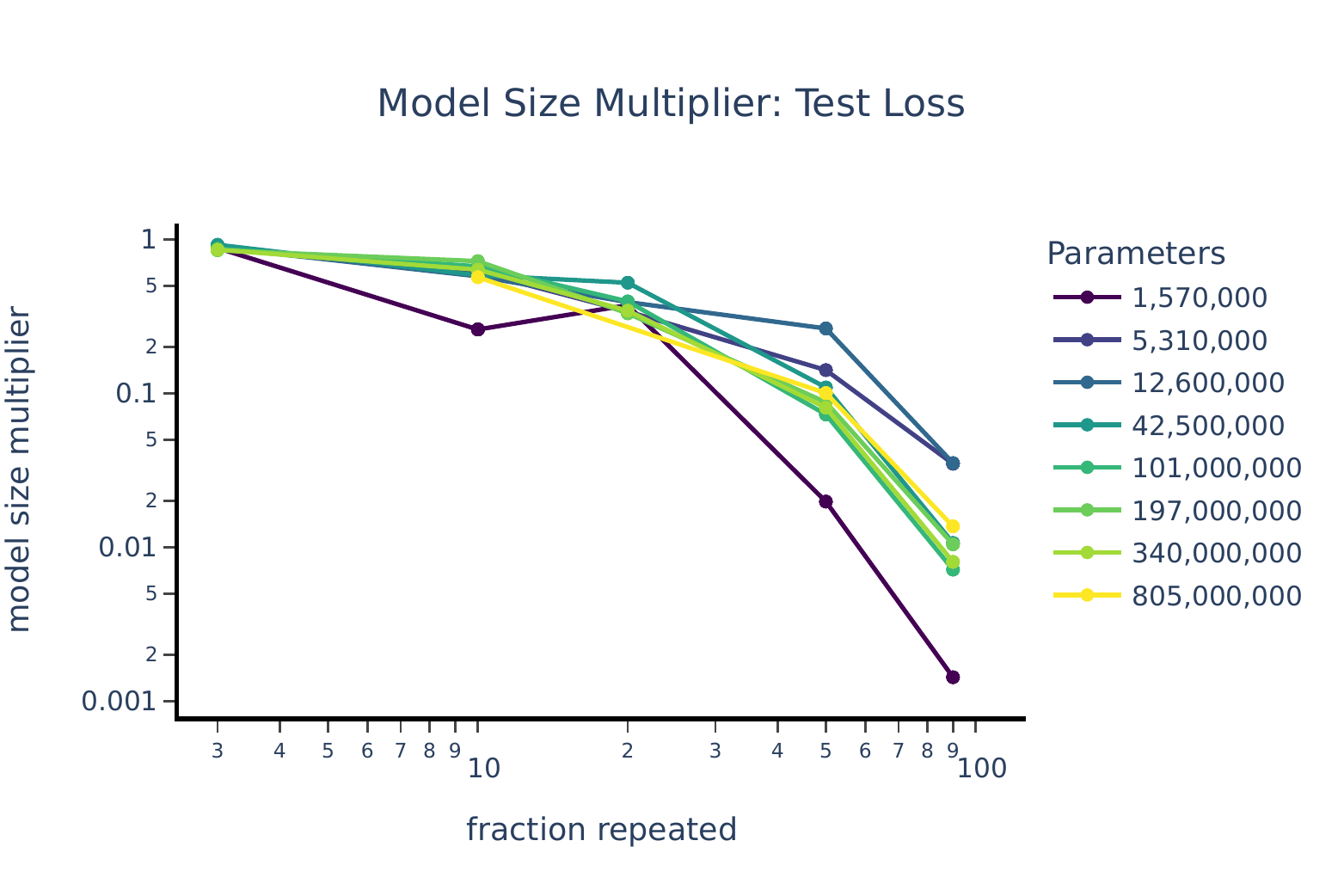}
    \caption{Comparison of degradation of prefix matching score with repeated data, compared to general degradation of the test loss.  We measured the double descent peak performance for a given model size and fraction of repeated data and compared that to a fit of the prefix matching score on the control model scan to generate an effective model size. We observe that 3\% repeated data causes on average \ref{fig:appendix_c_prefix} a 1.47 model size multiplier on prefix matching score while causing less than a 1.15x model size reduction in effective model size on test loss. Again we see much larger effects on the prefix matching score than on overall performance for repeated data fractions between 3\% and 20\%. The model size multiplier for prefix matching is based on a linear fit (see Appendix \ref{app:copy_prefix} for more details of fit). The test loss shown on the right is the same graph as in Figure 5, but with differently scaled axes for ease of comparison.}
    \label{fig:prefix_matching}
\end{figure}

We decided to probe the prefix matching score as measure of mechanistic structure that is distinct from the behavior of copying itself.  Figure \ref{fig:prefix_matching} shows the same setup as Figure \ref{fig:hp_copying_1} except for prefix matching score instead of copying loss.  As can be seen in the figure, preferential damage to prefix matching score is not present across the whole range of repeated data fraction as it is for copying, but at low fractions of data repeated, there is still preferential damage.  For example, at 3\% repeated tokens, there is a 2x effective parameter decrease in prefix matching score, but only a 1.15x effective parameter decrease in general (test) loss.

As another example, we find it interesting that the sharp drop in prefix matching score for a 1.5M parameter model with 50\% repetition corresponded to a complete breakdown of paragraph level copying.  This complete breakdown of paragraph level copying corresponds to a 1.5M parameter model having the effective overall performance of a 30,000 parameter model, while having an equivalent prefix matching score to a model with effectively 2,000 parameters.

Although not as conclusive as the previous results, these clearly show that prefix matching is preferentially degraded in some cases.

\begin{figure}
    \centering
    \includegraphics[width=0.49\textwidth]{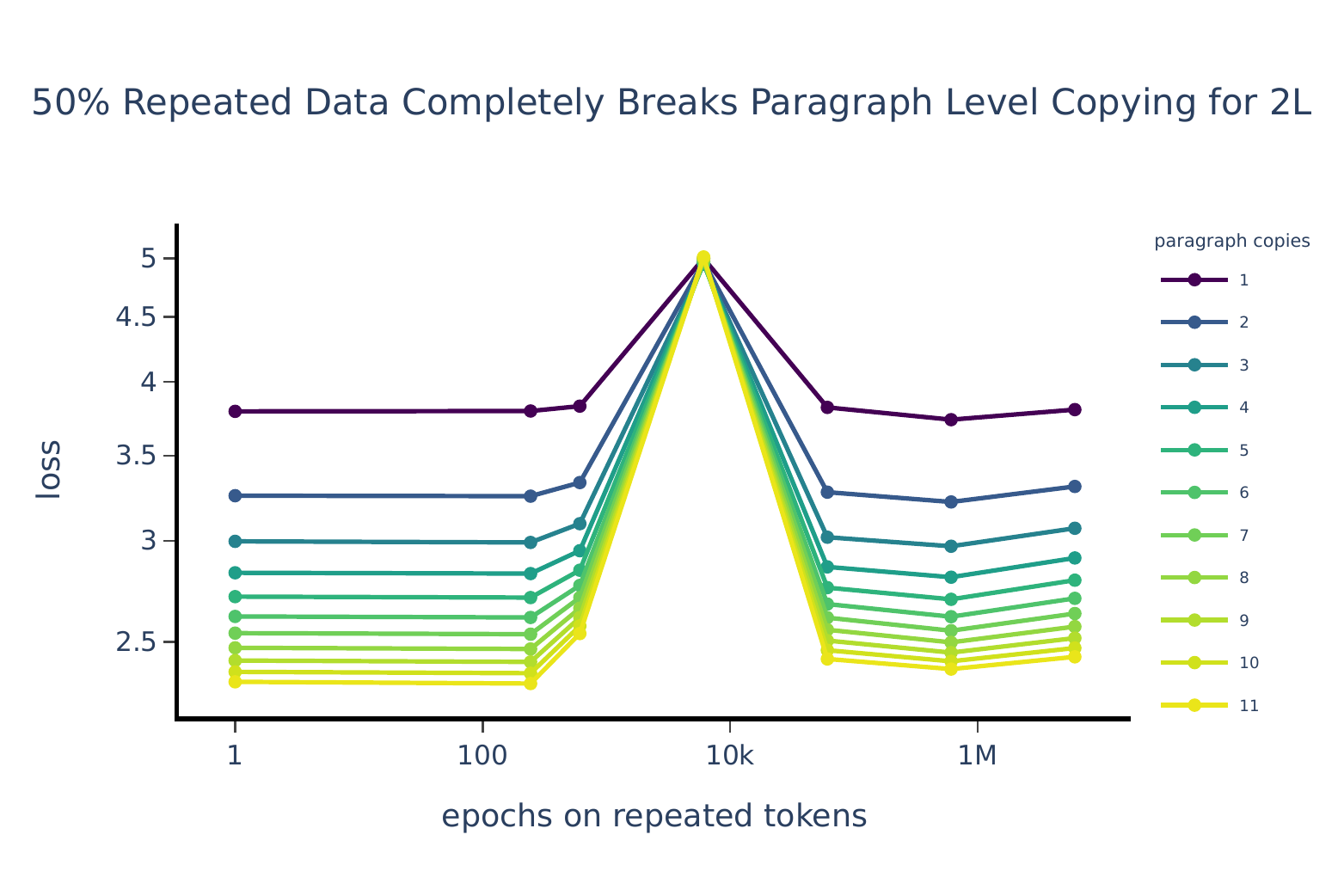}
    \includegraphics[width=0.49\textwidth]{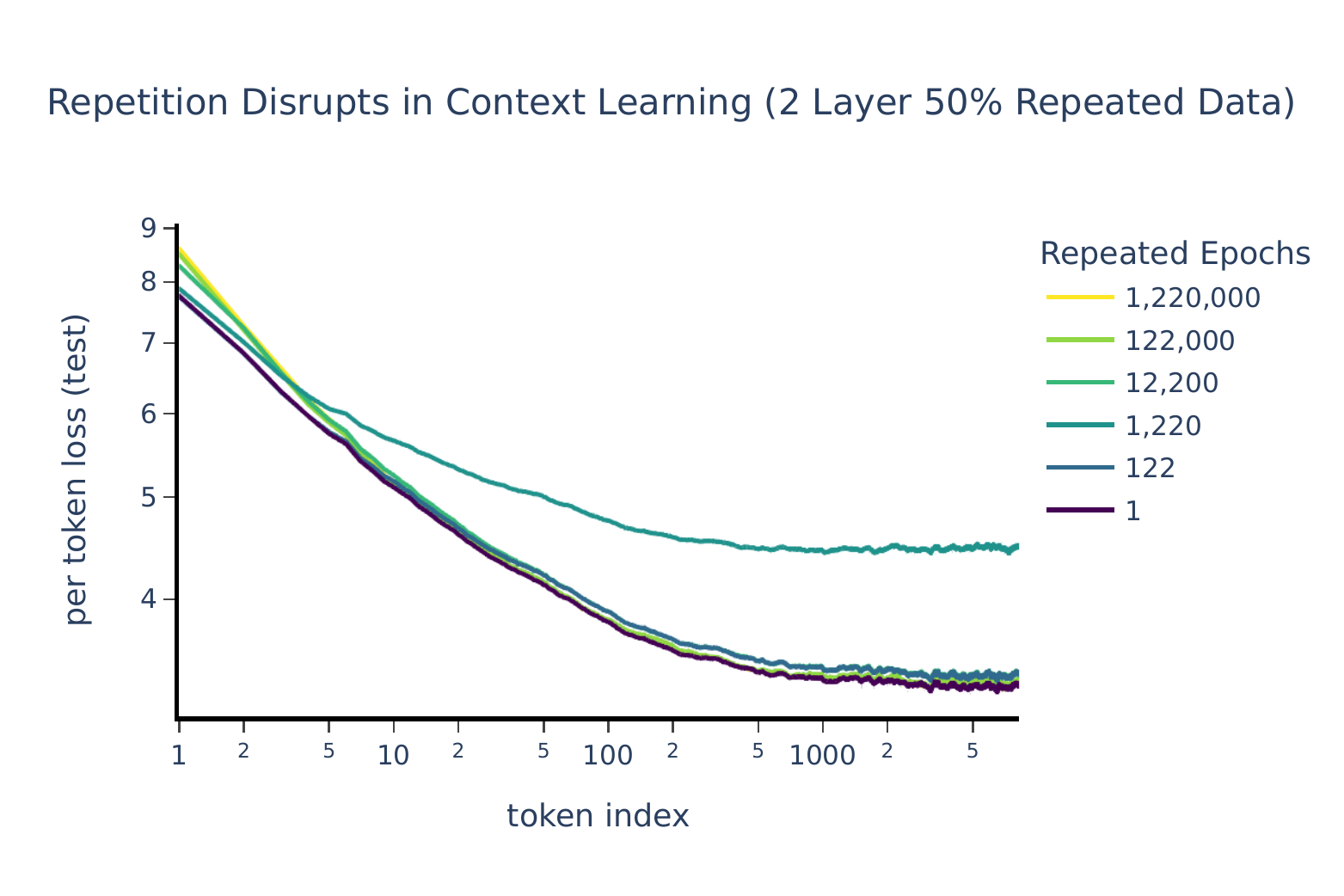}
    \caption{Degradation of copying and in-context learning at the peak of the double descent curve. On the left we show the 2-layer models trained on 50\% repeated data from Figure \ref{fig:hp_copying_1}, evaluated on the first paragraph of Harry Potter copied X times where X runs from 1 to 11. In Appendix \ref{app:hp_fewer_characters}, we explore shortening the length of the paragraph to verify the problem is with copying rather than long contexts. The right shows per token losses on the test set. Both graphs show dramatically reduced performance (higher copying loss, lower benefit to in-context learning) at the peak of the double descent.}
    \label{fig:hp_copying_2}
\end{figure}

\textbf{One and two-layer attention only models are worse at copying and fuzzily copying proper names on inspection}. To examine the effect on induction heads and in-context learning even more closely, we looked at more granular copying in one and two layer attention-only transformers, for which interpreting the internal structure (and especially induction heads) is known to be particularly straightforward \cite{elhage2021mathematical, olsson2022context}. That is, we can reverse engineer a large portion of attention-only-transformers (no MLP's) with a circuits-level understanding (understanding how individual neurons act together to produce useful behavior) \cite{cammarata2020thread:}. These small models also exhibit the same double-descent phenomenon as larger models (Appendix \ref{app:logit_attribution}).

\begin{figure}
    \centering
    \includegraphics[width=1.0\textwidth]{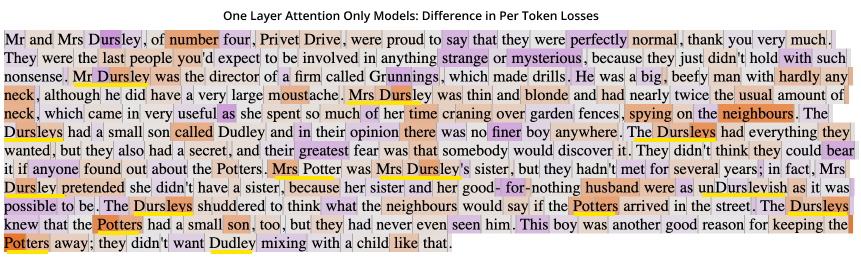}
    \caption{Visualization of the difference in loss on the first paragraph of Harry Potter for control and 10\%-repeated-data runs of a 1-layer attention-only model. Orange highlights correspond to the control model performing better, purple corresponds to the repeated data performing, and the intensity corresponds to the magnitude of the difference in per token losses.  Proper names (which are a good target for copying when they occur more than once) are underlined in yellow on second or later occurance; it is clear that the control model performs better on these. Often the difference is dramatic: for the last three appearances of ``Potters'' the control model puts a >97\% chance on ``ters'' given ``Pot'', whereas the repeated data model puts <4\% chance on that token.}
    \label{fig:1l_attn_hp}
\end{figure}

For 1-layer attention only models, where copying takes the form of skip-trigrams, we can easily see that the repeated data model is worse at a form of copying associated with these skip trigrams.  Namely, we compare the probabilities that the repeated data and control models assign to each token in a paragraph, and focus especially on proper names which occur repeatedly in the paragraph (Figure \ref{fig:1l_attn_hp}).  The most obvious way to correctly predict these re-occurring names is by copying, and we see that in most cases the control model (trained on unique text) performs much better than the one with repeated data (yellow underlines).  

Very specifically, predicting repeated names requires exactly a skip-trigram pattern \cite{elhage2021mathematical} which is the algorithmic operation 1-layer attention-only models are known to perform. For example, the following skip-trigrams are useful in the Harry Potter paragraph in Figure \ref{fig:1l_attn_hp}:

\[[a][b] \ldots [a] => [b] \;\;\;\;\;\;\;\; [\;Pot][ter] \ldots [\;Pot] => [ter] \]
\[[a][b] \ldots [a] => [b^\prime] \;\;\;\;\;\; 	[\;Pot][ter] \ldots [\;Pot] => [ters] \]
\newline

\begin{figure}
    \centering
    \includegraphics[width=1.0\textwidth]{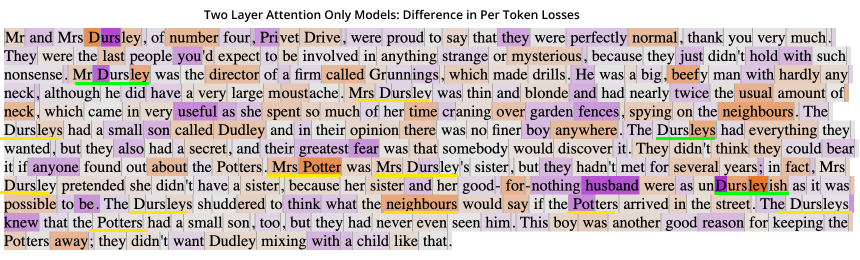}
    \caption{Same as Figure 9, but for 2-layer attention-only models. Proper names (which are a good target for copying when they occur more than once) are underlined in yellow on second or later occurance. Here the repeated-data model sometimes does better on repeated proper names, but there are still clear examples of the control performing much better. These examples are highlighted in green and discussed. On the token [ley] in the second appearance of [D][urs][ley] the control model places a 92\% likelihood on [ley] whereas the repeated data model places a 10\% likelihood. On the token [leys] in the second appearance of [D][urs][leys] the control model places a 44\% likelihood on [leys] whereas the repeated data model places a 4.9\% likelihood. On the [ley] in [ un][D][urs][ley][ish] the control model places a 68\% likelihood on [ley] whereas the repeated data model places a 0.4\% likelihood.
}
    \label{fig:2l_attn_hp}
\end{figure}

We also plotted the same visualization for a 2-layer attention-only model (which is known to contain simple induction heads), and find the control model is better at fuzzy copying (Figure \ref{fig:2l_attn_hp}).

Visually, it is less obvious (compared to the 1-layer case) that the 2-layer repeated model is worse at names, and there are a few examples where it puts 1.1x higher odds on the correct token. But on the other hand there are dramatic cases of the control model doing 500x times better (odds ratio on correct token) for fuzzy copying, like unDursleyish, which is exactly the kind of degradation we'd expect to see from disrupting induction heads. 

We attempted to leverage logit attribution (which earlier tokens contributed to the prediction of the current token through a "direct path" with this attention head) to see if the difference was primarily due to the induction head being less active or other heads interfering with it \cite{olsson2022context}. We were unable to find clear evidence of either, but we include our exploration of a 2 layer attention only model in Appendix \ref{app:logit_attribution}. 

\textbf{Repeated data causes a smaller, disproportionate performance drop on our out-of-distribution evaluations.}

\begin{figure}
    \centering
    \includegraphics[width=0.49\textwidth]{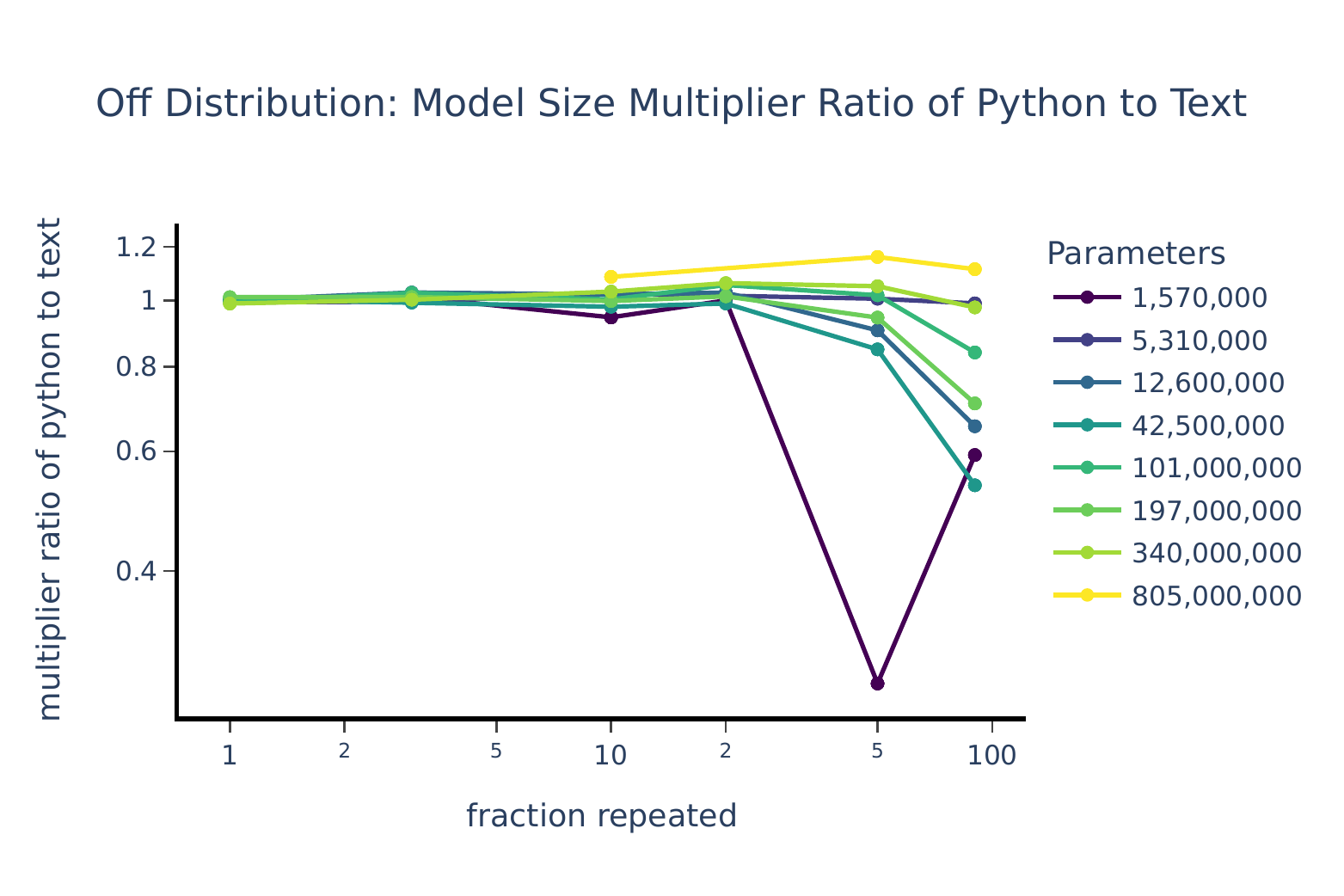}
    \includegraphics[width=0.49\textwidth]{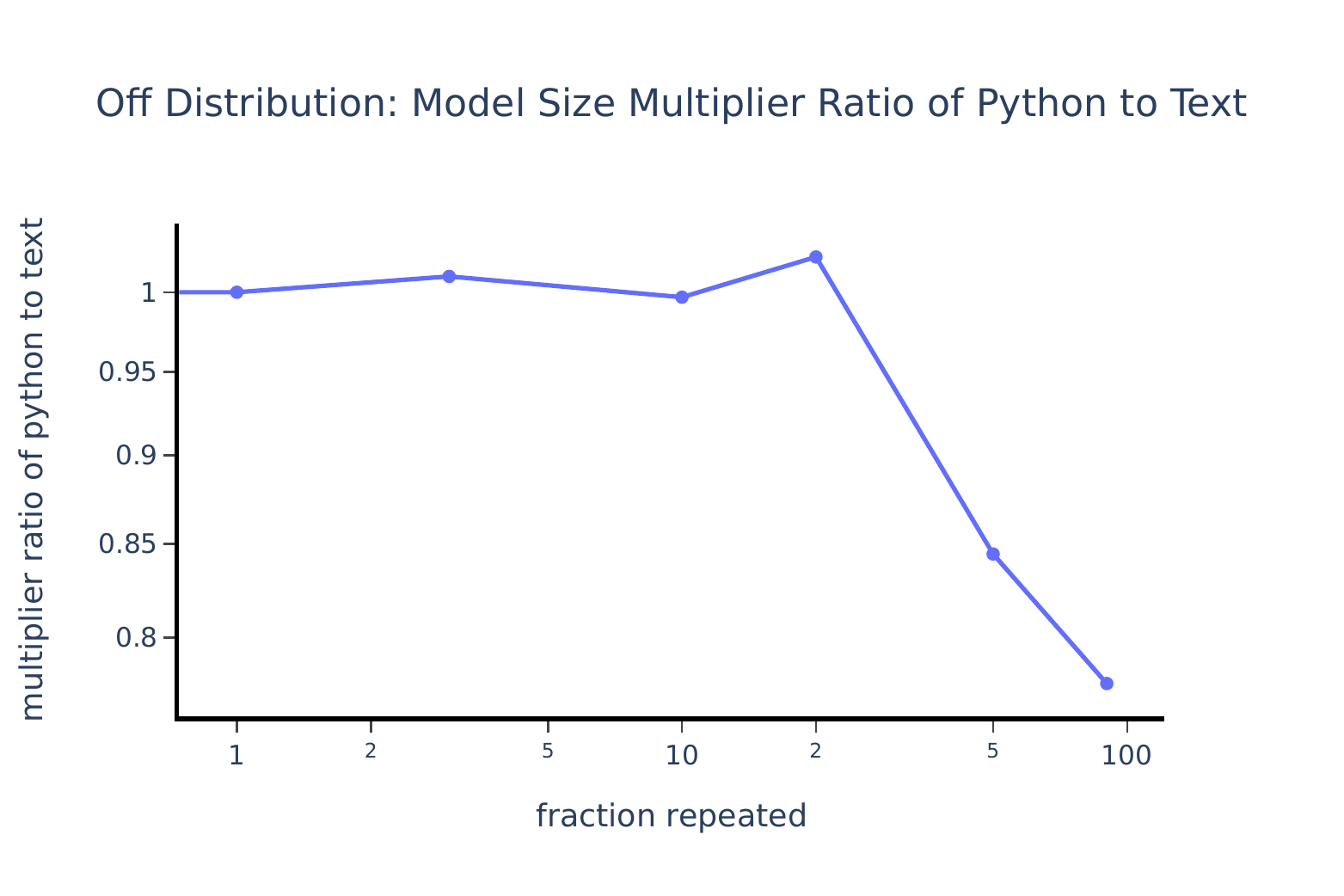}
    \caption{We observe that training on high levels of repeated data causes a small disproportionate drop on out-of-distribution performance (Python loss). The effect is noisy, but since we do not see a model size effect we take the average in the figure on the right (harmonic mean of multipliers). For large repeated fractions of 50\% and 90\% we see model size multipliers of .84 and .75.}
    \label{fig:ratio_py_text}
\end{figure}

Given that we overfit the model, we expected it to perform worse off distribution, which we do observe (Figure \ref{fig:ratio_py_text}). We notice almost an opposite pattern to what we observed in the induction head results. We see most of the disproportionate drop at 50\% and 90\% rather than 1-10\%. 

\begin{figure}
    \centering
    \includegraphics[width=.49\textwidth]{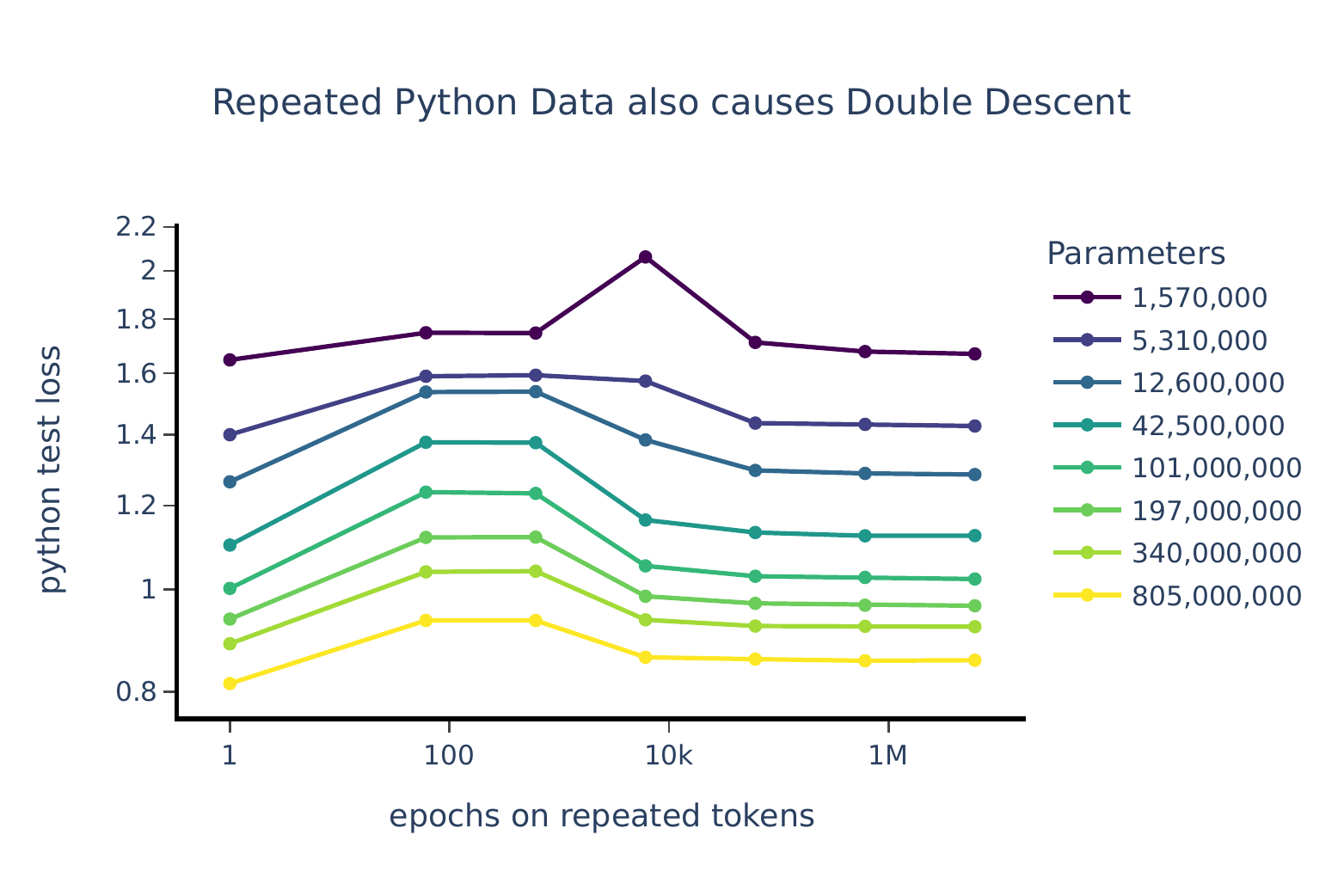}
    \includegraphics[width=.49\textwidth]{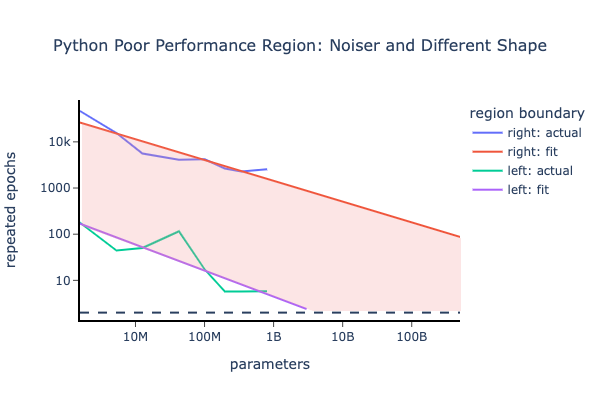}
    \caption{Double descent phenomenon for models trained on python. Training on Python gives similar results to what Figure 2 and Figure 4 show for language models.  Here 50\% of the dataset consists of repeats and 50\% is unique. On the left side is degradation in performance, occurring over a specific range of repetition that varies with model size.  On the right, we again see a large region of poor performance as we did in Figure \ref{fig:poor_scaling}, although the fit is noisier. Again the blue and green curves correspond to the right and left sides of the double descent curve where we observe 50\% of the maximum effect.}
    \label{fig:py_double_descent}
\end{figure}

\textbf{We observe a double descent phenomenon in sparse sweep of models trained on python, but we the Python scans exhibit a somewhat different overall shape}. To add more generality to our results, we repeated the same experiments on a Python dataset instead of natural language (Figure \ref{fig:py_double_descent}).  If we use the same method to fit the poor performance region, we see a broadly similar fit and a second epoch for today's large models (approximately 200B parameters) is still robustly in the reduced performance region for python. However the fit is noisier than the fit for text and the two lines are no longer parallel.

The noise may partially be explained by the Python fits being averaged over half as many settings for the fraction of tokens that are repeated data. It could also be that we need a higher resolution Python scan to get a cleaner estimate for the poor performance region. Finally, the Python data was trained on approximately 2 epochs as described in the methods section (so it included some repetition on the main dataset as well, not just the repeated subset). Python also may have more unintentional repetition than text, from copying and pasting of example code and forking of codebases. Such repetition could change the shape of the region of poor performance. More analysis of the Python experiments is shown in Appendix \ref{app:region_fits}.

\begin{figure}
    \centering
    \includegraphics[width=0.49\textwidth]{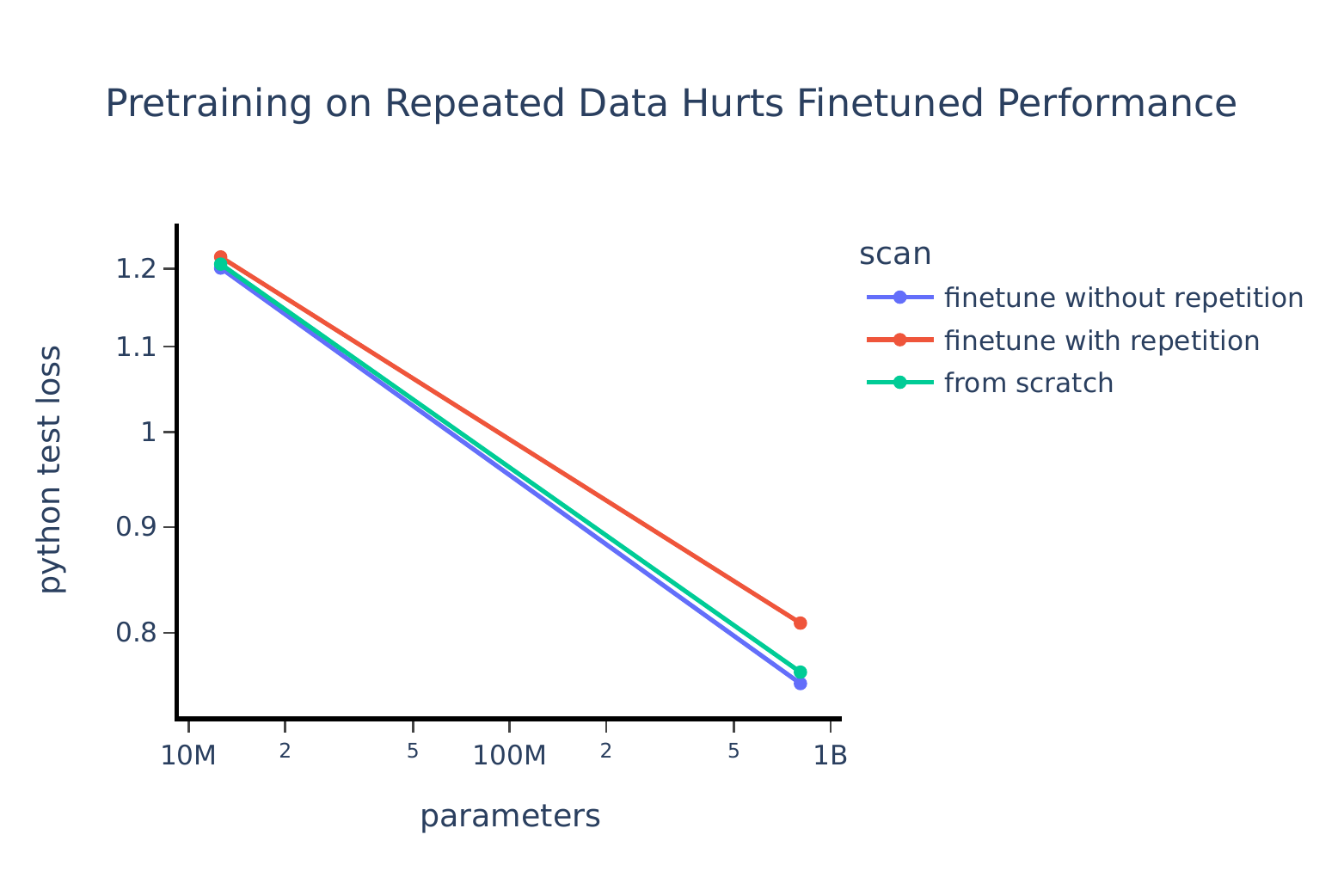}
    \includegraphics[width=0.49\textwidth]{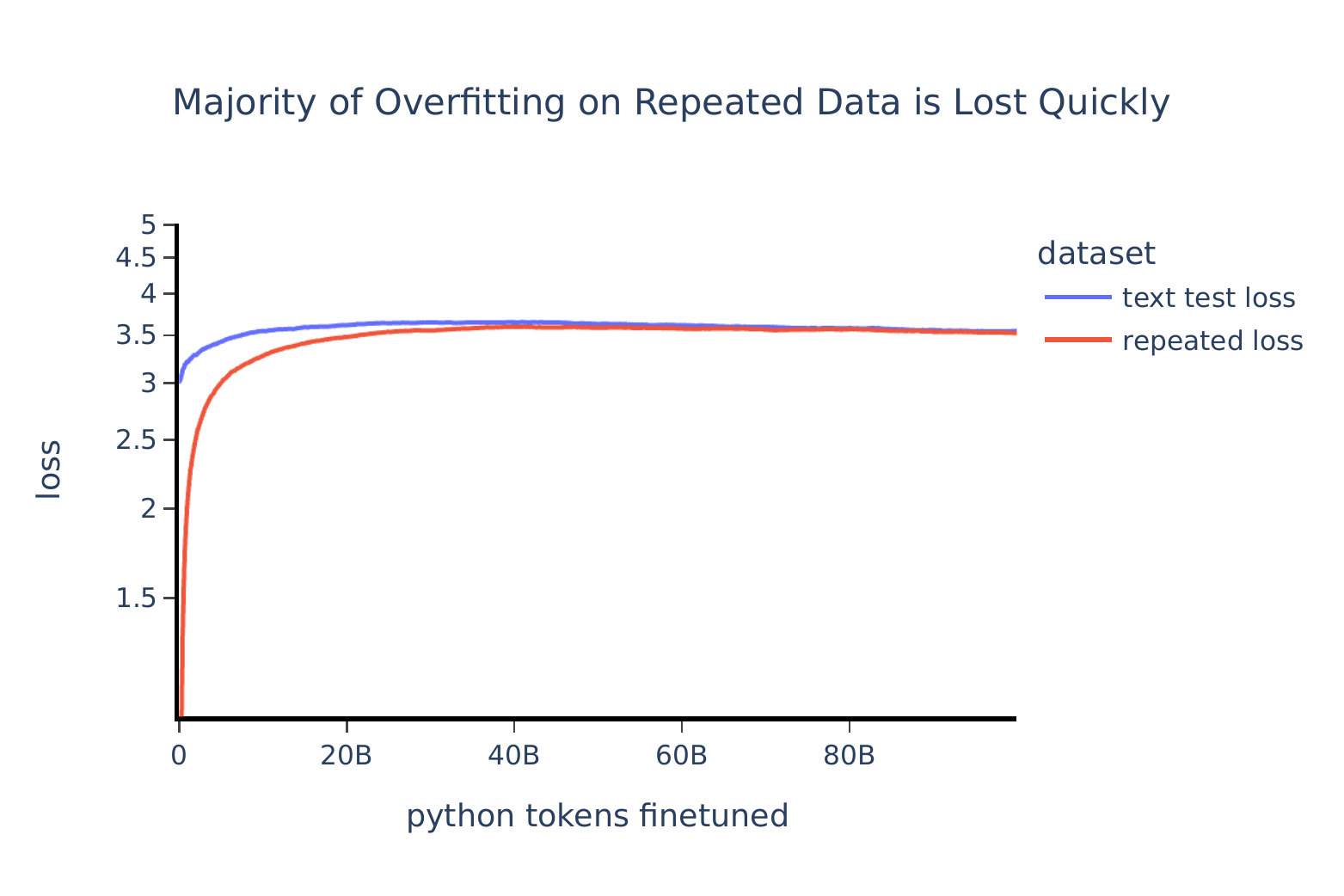}
    \caption{Effect of repeated data during pre-training on fine-tuning.  Models were pre-trained on 90\% repeated data (red lines) or on totally unique data (blue lines), and then fine-tuned on Python (always unique data).  The repetition frequency was chosen to maximize the performance hit.  The model pre-trained on repeated data encounters a sizable performance hit during fine-tuning (left panel), causing it to not only perform worse than the model pre-trained on unique data, but also worse than a model trained from scratch (green line).  The right panel shows fine-tuning curves of the two models.  The model pretrained on repeated data performs much worse for several billion tokens (red line), but eventually catches up to the model pretrained on unique data (blue line).}
    \label{fig:ossification}
\end{figure}

\textbf{Pre-training on repeated data hurts fine-tuned performance} We find that the negative impact of repeated data persists after fine-tuning natural-language models on Python (Figure \ref{fig:ossification}).

It is noteworthy that the performance hit once fine-tuned is much smaller. An 800M model pre-trained on 50\% repeated data from the double descent peak had its effective parameters reduced by 10x in Figure \ref{fig:1_noisy} in Appendix \ref{app:region_fits}. When we fine-tune from the repeated model we see a 1.6x reduction in effective parameters compared to training from scratch. This is still meaningful damage to the model, but it is recovered substantially. Since the repeated model forgets the repeated dataset after a modest amount of fine-tuning (Figure \ref{fig:ossification}, we consider the fine-tuned model with repeated data pre-training to be dominated by the fine-tuned model from the unique dataset.

\section{Methods}
The decoder-only transformer models were trained on an 8192 token context with the same settings as described in \cite{https://doi.org/10.48550/arxiv.2112.00861} for 100B tokens. Our language experiments utilized a 400B token dataset with 55\% heavily filtered common crawl data (220B tokens), 32\% internet books (128B tokens), and some smaller distributions including OpenWebText, Wikipedia, and Stack Exchange; most of which we sourced from The Pile \cite{https://doi.org/10.48550/arxiv.2101.00027}, and leveraged the 50,304 vocabulary GPT-2 encoding \cite{radford2019language, https://doi.org/10.48550/arxiv.1910.03771}. 

Code models were trained or fine-tuned on 45B tokens of Python for 2.2 epochs. Fine-tuning experiments had the same hyperparameters as pre-training experiments, but with learning rates reduced by a factor of 2 and reduced warmups.

We varied model size, repeated dataset size, and the fraction of tokens trained on repeated data by 3, 2.5, and 2 orders of magnitude respectively. 

\section{Related Work}
\textbf{Scaling Laws}
\newline
A scaling law lens consists of finding a small set of hyperparameters that have large, predictable impacts on model performance, and was present throughout this work (at least one of the hyperparameters is generally model size, compute, or dataset size). The predictive nature of scaling laws makes them useful in a broad number of research and engineering settings. The implications of scaling laws are sufficiently broad and understandable that understanding them is relevant to policy makers \cite{https://doi.org/10.48550/arxiv.2202.07785}. Predictable scaling trends in neural networks
were first studied with \cite{https://doi.org/10.48550/arxiv.1712.00409}. \cite{https://doi.org/10.48550/arxiv.2001.08361} demonstrated that test loss performance on language
modeling tasks scales as a predictable function of model size, dataset size, and compute. The scaling law lens has become more popular over time. For instance scaling
laws have been shown in many modalities (e.g., images, video, math, etc.) \cite{https://doi.org/10.48550/arxiv.2010.14701}, acoustics \cite{https://doi.org/10.48550/arxiv.2106.09488}, transfer to code,
\cite{hernandez2021scaling}, and few-shot adaptation of vision models \cite{https://doi.org/10.48550/arxiv.2110.06990}. Existing scaling laws have been revisited as training setups change; for instance, \cite{https://doi.org/10.48550/arxiv.2203.15556} found that many recent large models have been under-trained. Our work uses the scaling law lens on an aspect of dataset quality and supplements the lens with an interpretability lens, and we believe our work is novel in both these respects.

\textbf{Mechanistic Interpretability}
\newline
A mechanistic interpretability lens was used in this work. Mechanistic interpretability refers to attempting to reverse engineer the detailed computations performed by the model. The mechanistic interpretability lens is useful for pure scientific understanding and has the potential to anticipate safety issues from future more powerful models. There is a relatively detailed understanding of mechanistic interpretability for convolutional image models \cite{cammarata2020thread:}, some understanding for multimodal models \cite{goh2021multimodal, https://doi.org/10.48550/arxiv.2103.00020}, and such an understanding is starting to be built up for Transformers trained on language \cite{elhage2021mathematical, olsson2022context}. For a more thorough background on interpretability progress see the related work section of \cite{elhage2021mathematical}. These results are an example of a ``bridge'' between microscopic phenomena inside the network and macroscopic trends in the loss, and we're only aware of one other example of such a bridge \cite{olsson2022context}.

\textbf{Double Descent}
\newline
Double descent was first shown in generality by Belkin et al. \cite{https://doi.org/10.48550/arxiv.1812.11118} where it was observed for decision trees, random features, and 2-layer neural networks. Similar behavior has been observed in \cite{article, NIPS2001_26f5bd4a, https://doi.org/10.48550/arxiv.1710.03667, geiger2019jamming, https://doi.org/10.48550/arxiv.1912.02292}. For a more thorough background on double descent see Nakkiran et al. \cite{https://doi.org/10.48550/arxiv.1912.02292}. We extend the double descent phenomenon to a setting we see as more practical since data repetition in various forms appears to be a universal, long-term issue; whereas modern large language models are generally outside of the parameters and data regime of previously observed double descent phenomenon. 

\textbf{Rise of Engineering Large, Diverse Language Datasets}
\newline
Algorithmic innovation \cite{DBLP:journals/corr/abs-2005-04305}, compute \cite{amodei2018ai}, and data are three of the major factors that drive the advance of AI. The engineering and science of large, diverse language datasets is relatively new. Pre-2017 many language models were trained on a single distribution of text, such as news articles \cite{https://doi.org/10.48550/arxiv.1602.02410}, Wikipedia \cite{https://doi.org/10.48550/arxiv.1609.07843}, or fiction books \cite{https://doi.org/10.48550/arxiv.1506.06726}. GPT-2 \cite{radford2019language} leveraged webtext, outbound Reddit links with at least 3 upvotes in order to use human curation/filtration to ensure quality in addition to a broad distribution. GPT-2's capabilities are largely attributed to its scaled-up size and dataset (10x the parameters and 10x the data of GPT) \cite{radford2019language}. The next generation of language models, \cite{https://doi.org/10.48550/arxiv.2005.14165, https://doi.org/10.48550/arxiv.2112.11446, https://doi.org/10.48550/arxiv.2203.15556}, leveraged large, diverse datasets that consist of many sub-distributions. Constructing such datasets includes a large number of decisions: choosing sampling weights, quality filtering, de-duplication, fuzzy de-duplication, epochs per dataset, and more. There has not yet been substantial public work that quantitatively shows the impact of such decisions, but the dataset ablations in Appendix A of the Gopher \cite{https://doi.org/10.48550/arxiv.2112.11446} paper are notable. They clearly show the benefit of their dataset mixture, quality filter, exact de-duplication, and fuzzy de-duplication for 1.4B parameter models. Our work aims to provide some insights and potential diagnostics for researchers and engineers designing large datasets for language models.

\section{Discussion}
\label{discussion}
\subsection{Why does repeating a small fraction of data damage performance so much?}
We showed that a dataset with only 10\% repeated tokens can reduce model performance by an effective 2x in parameter count, much more than if that 10\% of the data had simply never been trained on. The repeated data thus degrades model performance out of proportion to its share in the dataset. Why does this occur, and why only for a specific amount of repetition?  One plausible hypothesis comes from looking at the model's ``incentives'' to memorize vs generalize. To informally explore this hypothesis consider the following rough numbers, a 800M parameter model typically has a loss of roughly 2.0 nats/token, a 400M parameter model has a loss of roughly 2.2 nats/token, and fully memorized data will have a loss of 0 nats/token. Now suppose a 800M model is trained on 90\% unique data and 10\% tokens consisting of repeated data. We can ask whether it is a ``good tradeoff'' for the model to memorize the repeated data (leading to 0 loss on 10\% of the dataset), at the cost of degrading performance by the equivalent of a 2x multiple in model size (which raises loss on the other 90\% from 2 to 2.2). Some simple arithmetic suggests that it is: $0.9 * 2.2 + 0.1 * 0 = 1.98 < 2.0$. Another way to say this is that zero loss is such a huge drop compared to the differences in entropy between model sizes that driving the loss to zero on even a tiny subset can incentivize enormous degradation in quality.

This however leaves open the question of when this tradeoff is necessary or possible -- and here is where the double descent phenomenon comes in. If a lot of data is repeated only a few times (say 5\% of the data repeated 2x) then the model may not have the capacity to memorize it, and also does not see it enough times during training to do so. If a tiny amount of data is repeated very many times (say 0.01\% of the data repeated 1000x), then the model will memorize it, but because it is so small the model need not use much capacity to do so, so the degradation in quality will likely be small. There is a range in the middle where the data can be memorized \it and\rm\space doing so consumes a large fraction of the model's capacity, and this may be where the peak of degradation occurs.

\subsection{Generalization, memorization, and induction heads}
Our results show that overfitting on the repeated data results in worse test loss, and this co-occurs with a disproportionate degradation in the model's induction heads (prefix matching score) and its ability to copy text. Copying sequences can be seen as a form of generalization, as it requires algorithmic operations that are independent of the content of the data. \cite{elhage2021mathematical, olsson2022context} provided evidence for induction heads as the mechanism implementing copying and other pattern-matching. For the 2 layer model shown in Figure \ref{fig:hp_copying_2} it seems as if the pressure to memorize the repeated dataset has led a skip tri-gram head to replace the induction head entirely. Thus our results tell a story where a type of generalization and its internal implementation are disrupted when the model memorizes repeated data -- a vivid illustration of the memorization-generalization trade-off. Future work could take this even further, by measuring the number of parameters devoted to memorization and trying to observe them competing for space with induction heads. Finally, it is worth noting that the co-occurence of copying degradation and induction head degradation is itself some additional evidence for induction heads as the source of in-context learning; Olsson et al. \cite{olsson2022context} was not fully conclusive and our results further bolster the case.

\subsection{Bridging mechanistic interpretability and scaling laws}
The results connecting memorization to the degradation of mechanistic interpretability structures \cite{olsson2022context} are an example of a ``bridge'' between microscopic phenomena inside the network and macroscopic trends in the loss. We view such connections as very fruitful tools for research, because they allow us to see the same thing through different lenses: the macroscopic behavior demonstrates the significance of the microscopic mechanisms, and the microscopic mechanisms help explain how and why the macroscopic phenomena occur. Switching back and forth between the two allows for a deeper understanding of both, as well as more robust diagnostics if something goes wrong. We are aware of at least one other instance of such a bridge -- the correspondence between the formation of induction heads and the boost in in-context learning near the beginning of training \cite{elhage2021mathematical, olsson2022context} -- but such connections remain rare so far, and we believe that finding more of them is a promising route to more deeply understanding neural nets.


\subsection{Repeated data and fine-tuning} We hypothesized repetition might help explain why models trained from scratch sometimes outperformed models that were pre-trained and then fine-tuned \cite{hernandez2021scaling}. For our purposes, we define ossification as any pre-training that leads a fine-tuned model to perform worse than a model trained from scratch (given a fixed compute and data budget). It required relatively extreme repetition in pre-training (90\% training on repeated tokens at  peak of double descent curve, 73x reduction in effective model size) to see a large ossification effect (1.6x reduction in effective model size) within our fine-tuning setup. We still think repetition might explain a large fraction of ossification when we consider training on various types of repetition we did not study here (sentence level, paragraph level, similar documents, distribution, etc). Overall, our finding that repetition can induce ossification provides medium causal evidence to this hypothesis. We think ossification is an interesting phenomenon that merits further study.

\subsection{Limitations}
We attempt to discuss limitations throughout the text where appropriate, but for the reader's convenience, we enumerate them here. We attempt to list them in a loosely descending order of importance.
\begin{enumerate}
    \item We used a fixed number of tokens for all models (similar to the GPT-3 model sweep), because these models were trained prior to the release of Chinchilla, which showed the compute frontier (pareto frontier of performance and compute) is quite different than previously understood \cite{https://doi.org/10.48550/arxiv.2005.14165, https://doi.org/10.48550/arxiv.2203.15556}. 
    \item Our fits for region of poor performance were relatively noisy, and we only observed a clean trend by aggregating them. This is discussed in the Results section and further explored in Appendix \ref{app:region_fits}. 
    \item The data we repeated was a random subset of the original dataset, and is thus not directly applicable to the situation where higher quality data (such as Wikipedia) is intentionally repeated to improve quality. Nevertheless, it seems plausible that the results would carry over.
    \item We measured loss, rather than downstream NLP evaluations. Overfitting does not always entail worse performance on downstream tasks \cite{https://doi.org/10.48550/arxiv.2203.02155}, so it is possible that the degradation we observe does not carry over to these tasks.
    \item We did not explore the effects of early stopping, dropout, weight decay, or other regularization.
    \item We did not investigate simpler systems than 1L attention-only models, which might contain more complete mechanistic insights.
\end{enumerate}

\subsection{Future Directions}
Below are some future directions we think are promising:
\begin{enumerate}
    \item A compute efficient frontier scan to predict the poor performance region.
    \item Varying the type of repetition. We could inject repeated sentences or paragraphs at the beginning or end of some fraction of contexts, or repeat chunks of documents in a different order. We could also explore cases where the repeated data has a different distribution than the unique data.
    \item Further interpretability work. Are there neurons that tell the model what distribution it is in: unique or repeated? Are there neurons through which we can observe and edit the repeated sequences?
    \item Drill down on memorization and generalization. Could we measure the number of model parameters taken up by memorization vs generalization, either behaviorally or by using mechanistic interpretability to identify parameters that are storing memorized data? Can we measure how this varies across the double descent, and thus watch the competition between memorized data and induction heads for model capacity?
    \item Could repetition and double descent help explain loss spikes during training? If a model can largely memorize a particularly easy batch in a single gradient step then a very skinny double descent could present as a loss spike.
\end{enumerate}

\section{Conclusion}
We've shown that small fractions of repeated data, if repeated at the right frequency, can cause surprisingly severe degradation to model performance. We show that this degradation scales predictably, occurs across datasets, and is associated with disproprotionate damage to internal mechanisms associated with generalization, such as induction heads. In practical terms, these results provide a tool for predicting and diagnosing data-repetition-related problems in language models. In more conceptual terms, they are an example of a bridge between the macroscopic domain of scaling laws and the microscopic domain of mechanistic interpretability, as well as a lens for gaining a more detailed understanding of how generalization and memorization work. We believe these conceptual themes are promising ones, and hope to see more work that employs them.

\section*{Acknowledgments}

We thank Ethan Perez, Jan Leike, and Martin Wattenberg for helpful feedback on the draft. We thank Daniela Amodei, Jamie Kerr, Jia Yuan Loke,  Rebecca Raible, and Tim Telleen-Lawton for support with the project.

\section*{Author Contributions}

\textbf{Danny Hernandez} led the project performed the majority of experiments, analysis, and writing.

\textbf{Tom Brown} led engineering efforts for the scaling team, including efficient pre-training and gave helpful feedback on the paper.

\textbf{Tom Conerly} made engineering contributions on the scaling team.

\textbf{Nova DasSarma} managed the underlying cluster infrastructure.

\textbf{Dawn Drain} helped with pre-training research and infrastructure.

\textbf{Sheer El-Showk} helped with pretraining research and dataset construction.

\textbf{Nelson Elhage} contributed significantly to interpretability tooling, provided support on that tooling, and gave helpful feedback.

\textbf{Zac Hatfield-Dodds} helped with codebase maintenance and with engineering

\textbf{Tom Henighan} helped with pretraining the underlying language models, with dataset creation, with managing the cluster during some phases of the project, and gave helpful feedback on the paper.

\textbf{Tristan Hume} contributed to interpretability tooling that was leveraged in this work.

\textbf{Scott Johnston} helped with pretraining research.

\textbf{Ben Mann} contributed to pretraining and cluster management.

\textbf{Chris Olah} lead the interpretability team, which provided tooling and support for this work.

\textbf{Catherine Olsson} contributed to interpretability tooling, provided support on that tooling, and provided interpretability research advice.

\textbf{Dario Amodei} contributed greatly to the framing and writing of the work and advised the project.

\textbf{Nicholas Joseph} helped design and build a framework for efficient training of large language models, gave helpful feedback on the paper, and advised the project.

\textbf{Jared Kaplan} led pre-training efforts initially and advised the project.

\textbf{Sam McCandlish} led pre-training efforts and advised the project.

\appendix
\addtocontents{toc}{\protect\setcounter{tocdepth}{1}}

\newpage
\section{Model Size Multiplier and Poor Performance Region Fits}
\label{app:region_fits}

\begin{figure}
    \centering
    \includegraphics[width=0.49\textwidth]{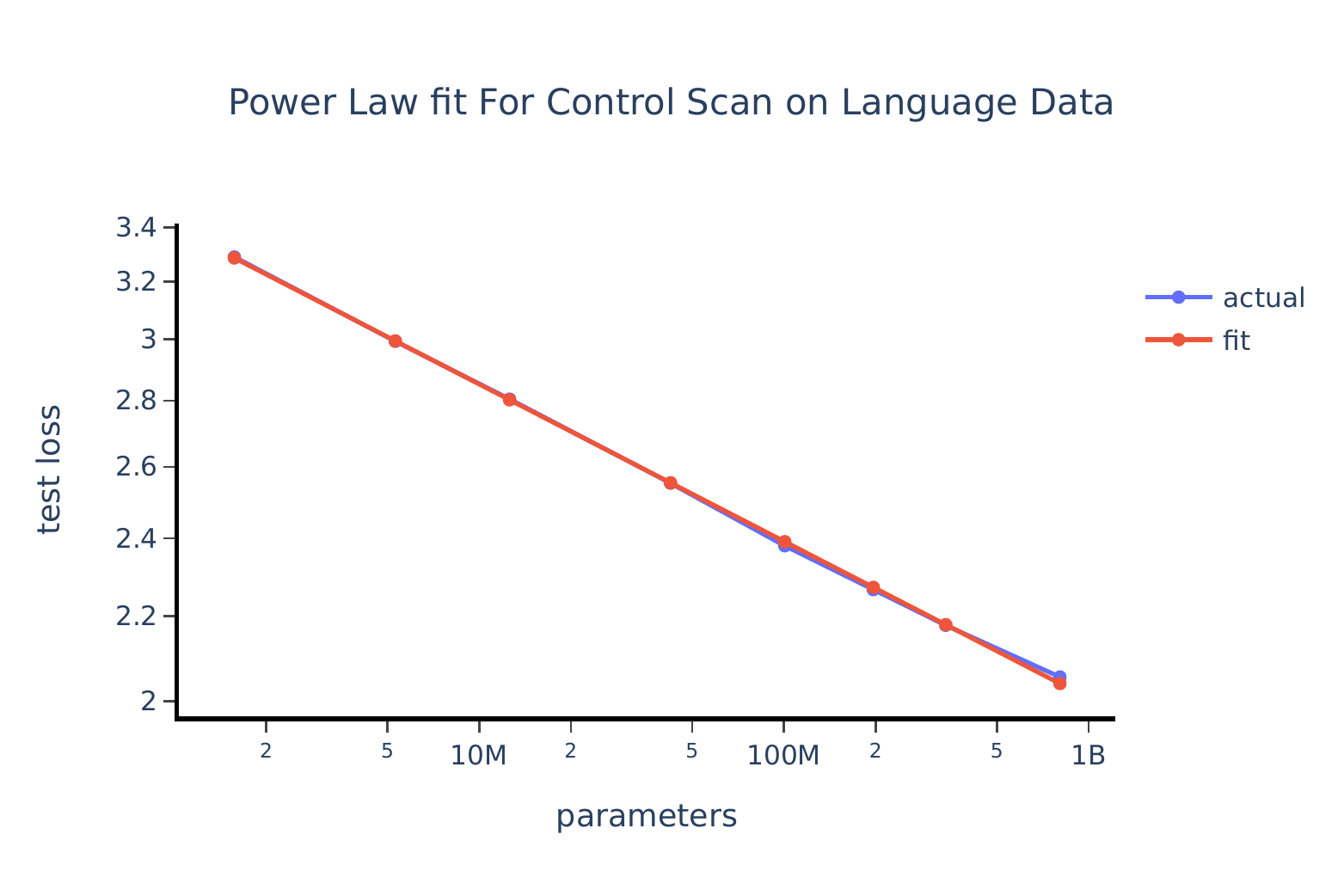}
    \includegraphics[width=0.49\textwidth]{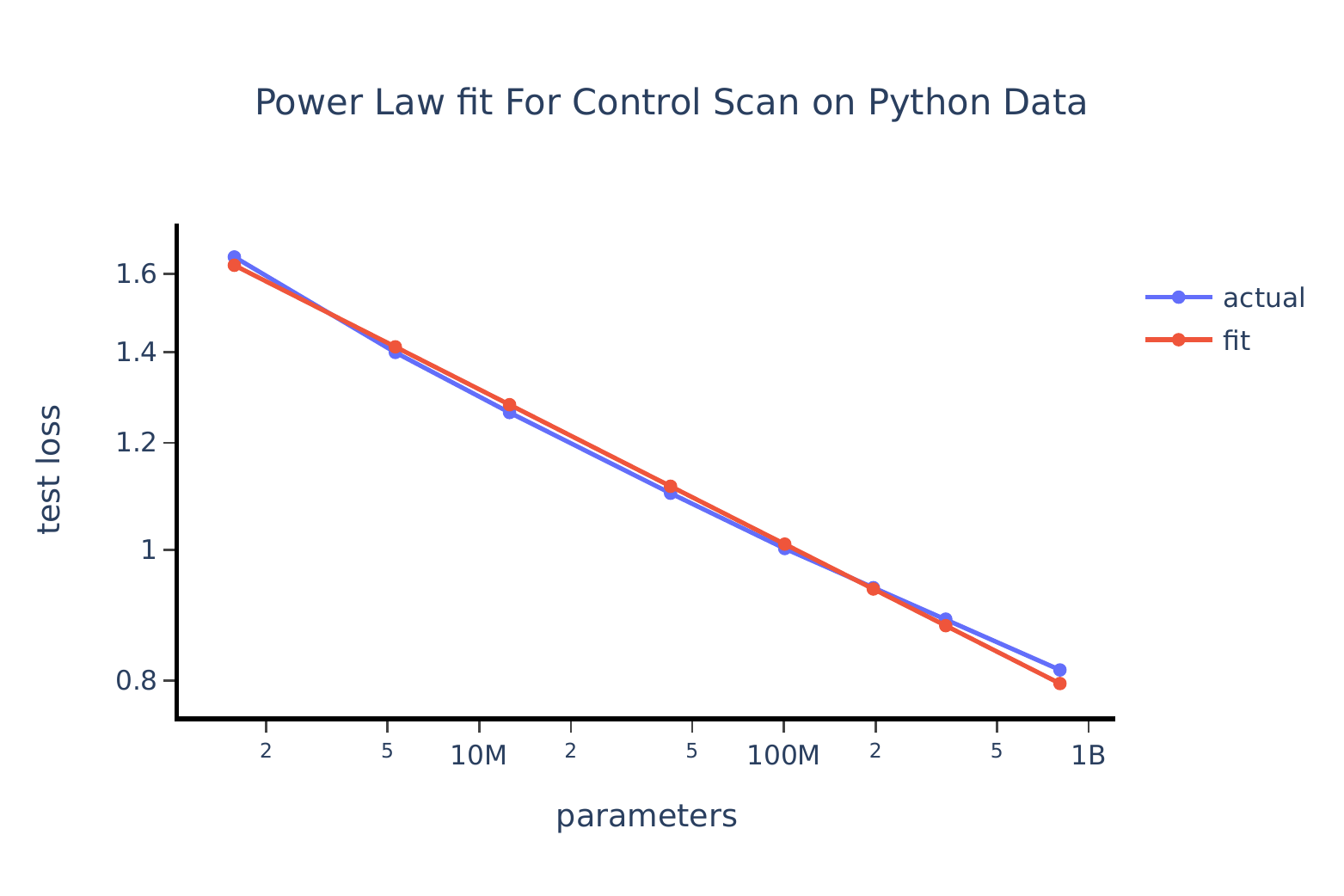}
    \caption{We see a power laws provide good fits for both language and Python data. We can use these fit to re-parameterize loss for our models trained on repeated data into model size multipliers.}
    \label{fig:power_law_fits}
\end{figure}

In order to fit the poor performance regions we first fit power laws to our control scans on language and Python so that we can re-parameterize loss in terms of model size multipliers. These fits are shown in Figure \ref{fig:power_law_fits}

When we graph repeated epochs vs model size multiplier with a given fraction of repeated data in Figure \ref{fig:1_noisy}, we observed that our 1\% repeated data graphs were quite noisy, so we excluded the 1\% scans from the fits. The 3\% repeated data graphs looked reasonable, in that the double descent peak looked large compared to the noise, so we included that all higher fractions in our fits.

\begin{figure}[h]
    \centering
    \includegraphics[width=0.49\textwidth]{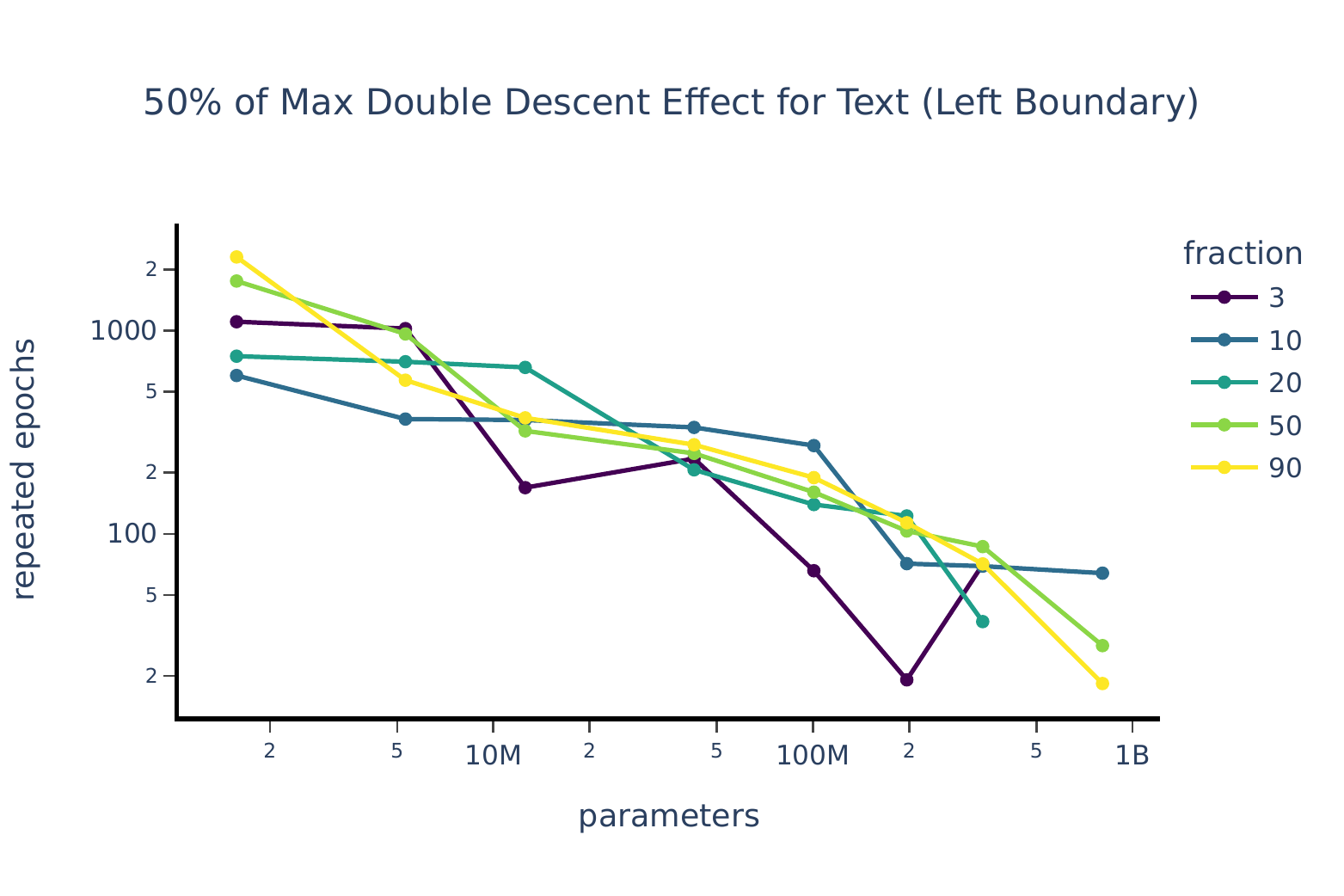}
    \includegraphics[width=0.49\textwidth]{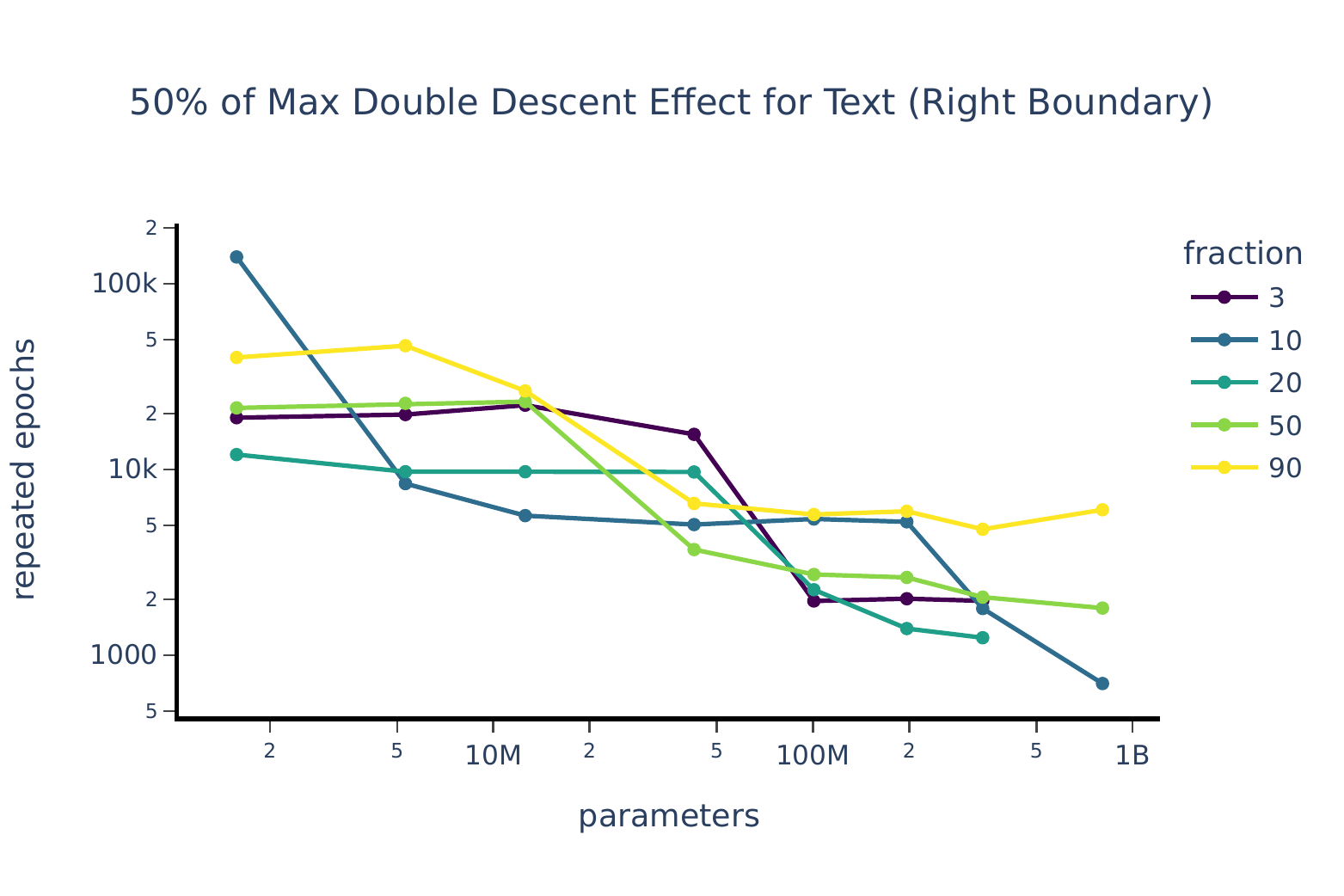}
    \caption{We estimate how many repeated epochs half of the maximum effect size (on a log scale) would be observed using linear interpolation on the left and right side of the double descent peak for each fraction of repeated data. We then averaged these curves to make an overall estimate for the left and right boundaries of the poor performance region shown in Figure \ref{fig:poor_scaling}}
    \label{fig:text_region_fitting}
\end{figure}

\begin{figure}
    \centering
    \includegraphics[width=0.49\textwidth]{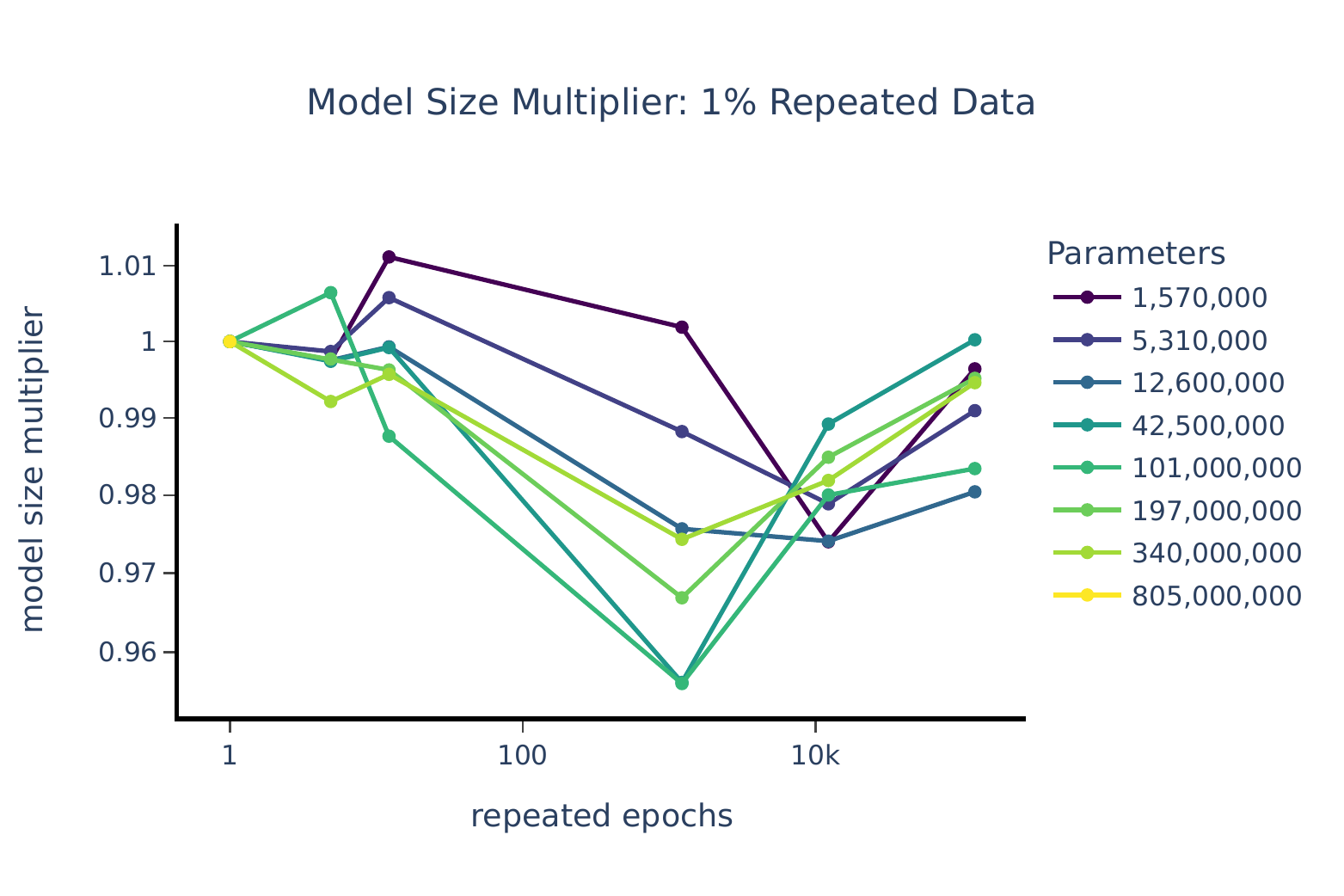}
    \includegraphics[width=0.49\textwidth]{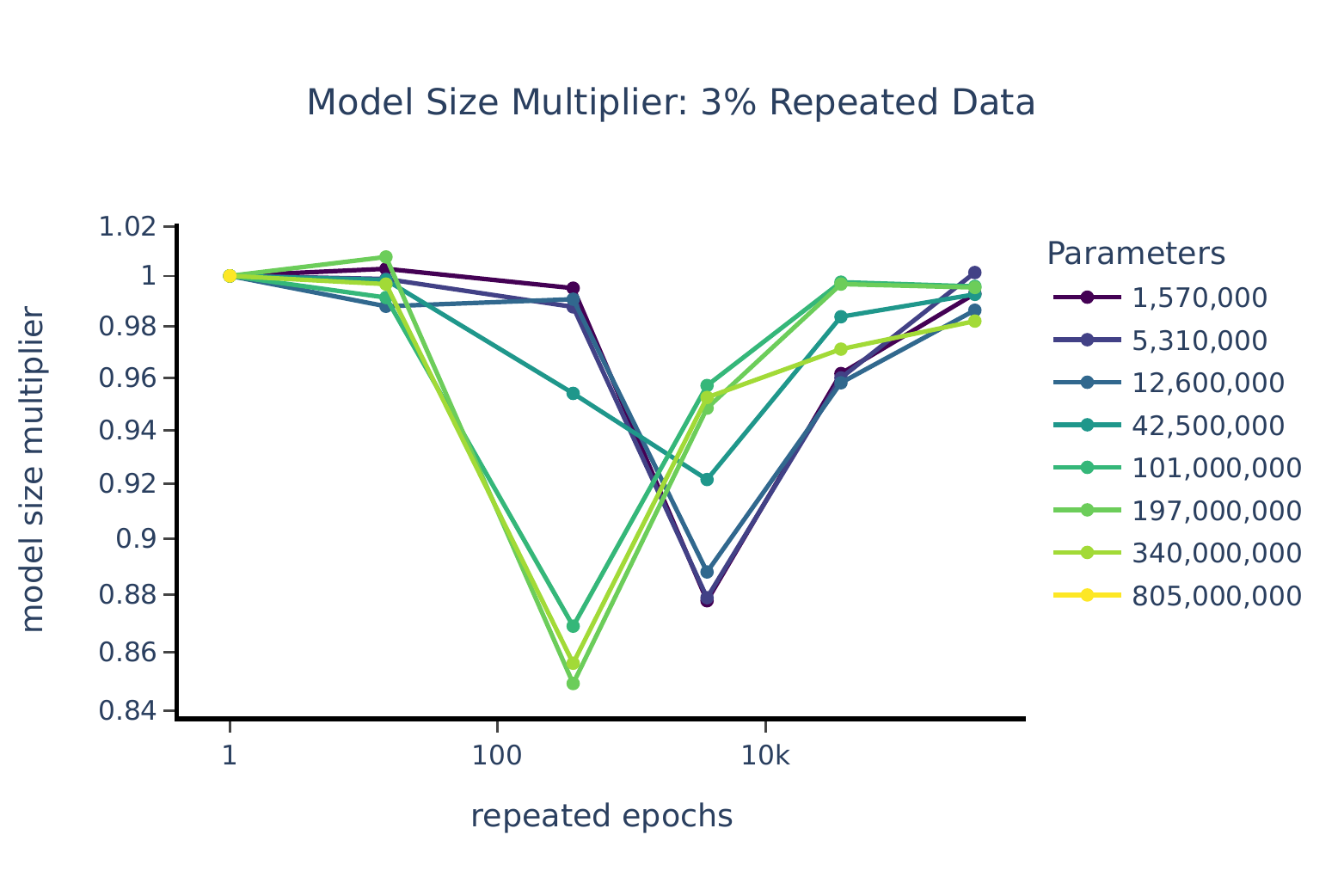}
    \includegraphics[width=0.49\textwidth]{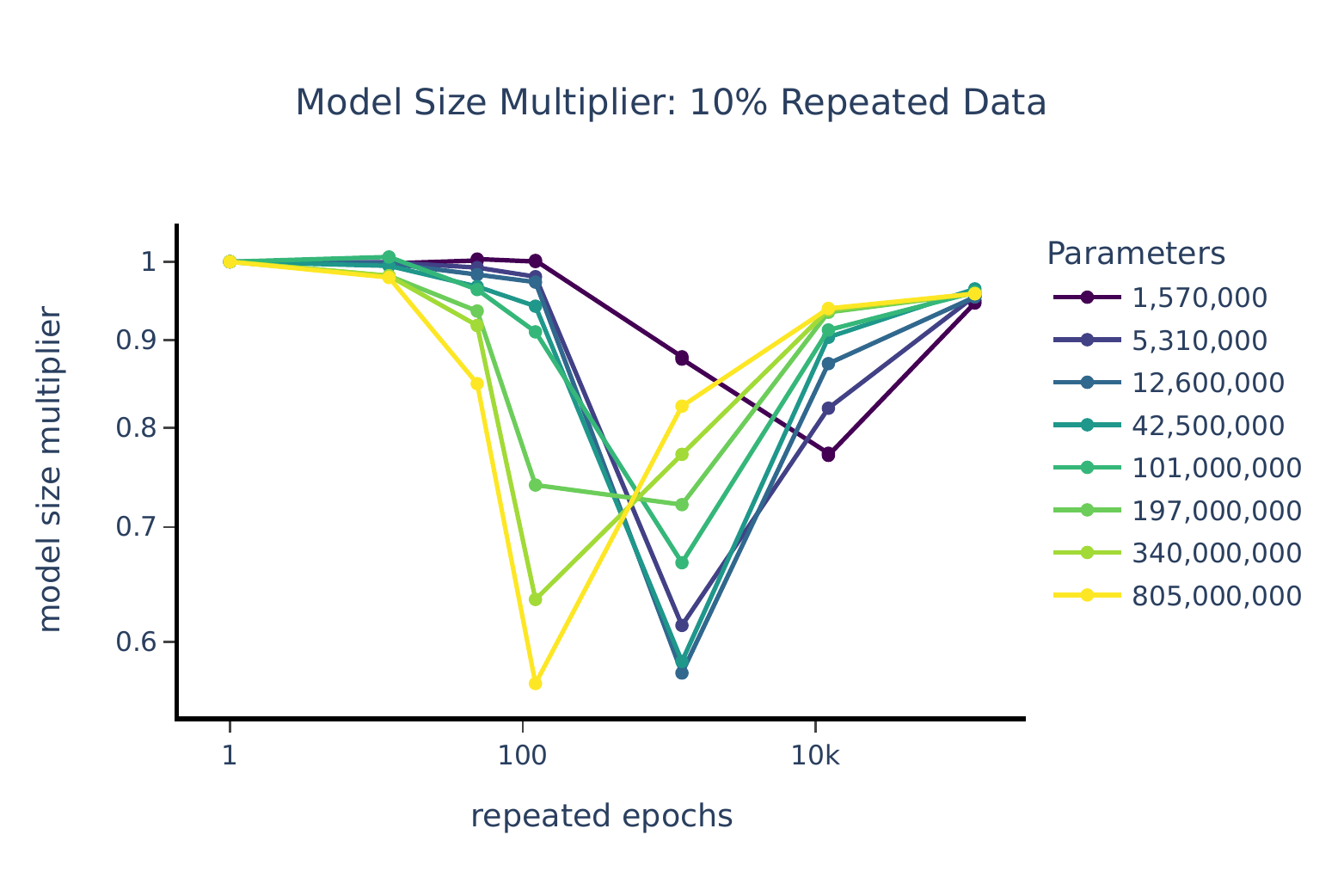}
    \includegraphics[width=0.49\textwidth]{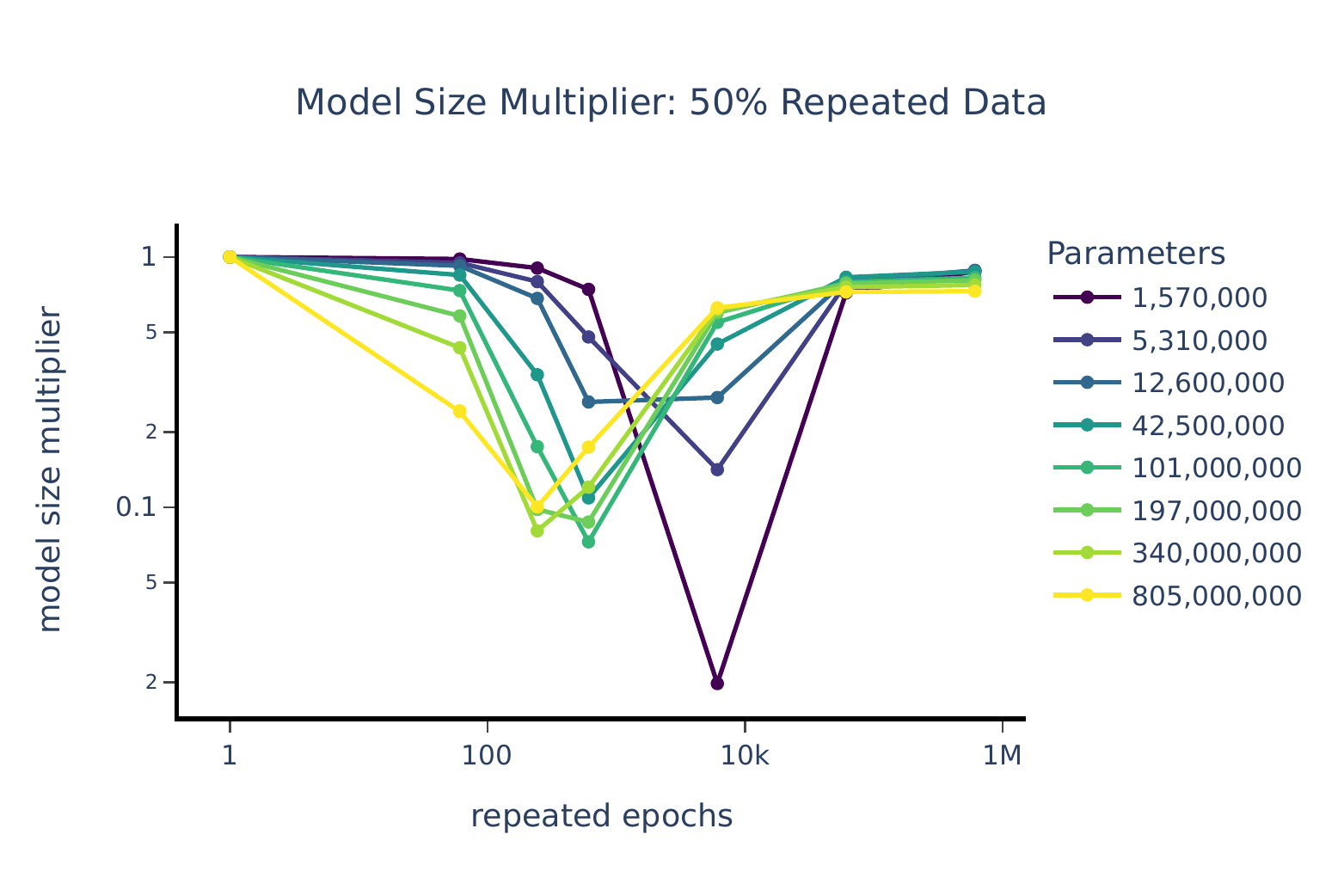}
    \caption{it is easier to see the sharpness of the double descent peaks in this diagram than Figure \ref{fig:double_descent}. The 1\% runs was much noisier than the rest so we excluded it from our fits}.
    \label{fig:1_noisy}
\end{figure}

We estimate how many repeated epochs half of the maximum effect size (on a log scale) would be observed using linear interpolation on the left and right side of the double descent peak for each fraction of repeated data. We then averaged these curves to make an overall estimate for the left and right boundaries of the poor performance region shown in Figure \ref{fig:poor_scaling} and Figure \ref{fig:py_double_descent}. For text this produces a relatively clean overall fit, but the the individual curves for text are relatively noisy as shown in \ref{fig:text_region_fitting}. Some potential explanations for the noise are i) Given the resolution of our scan we do not always get a good estimate of the peak effect for a given curve (the peak can easily be between two points we measured ii) our linear interpolation also introduces error as our underlying curves only have 6 points.

Overall we think the region of poor performances we showed in Figure \ref{fig:poor_scaling} is relatively robust in that it is useful to think about the sub distribution double descent phenomena there. However, we would not claim that we have produced extremely accurate estimates for the exact boundaries, even in our setup, and the boundaries could vary meaningfully given a different setup, especially differences in regularization.

\begin{figure}
    \centering
    \includegraphics[width=0.49\textwidth]{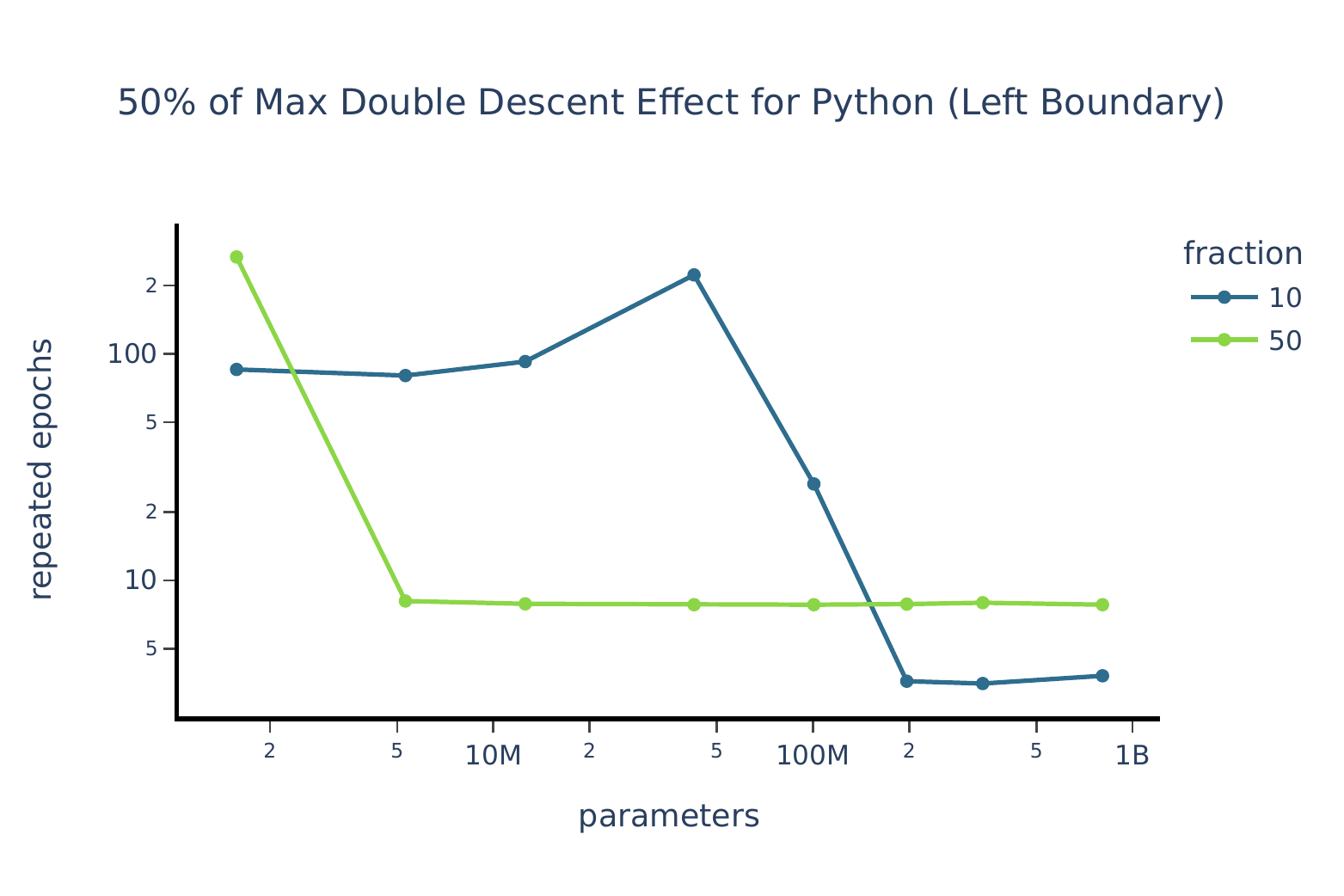}
    \includegraphics[width=0.49\textwidth]{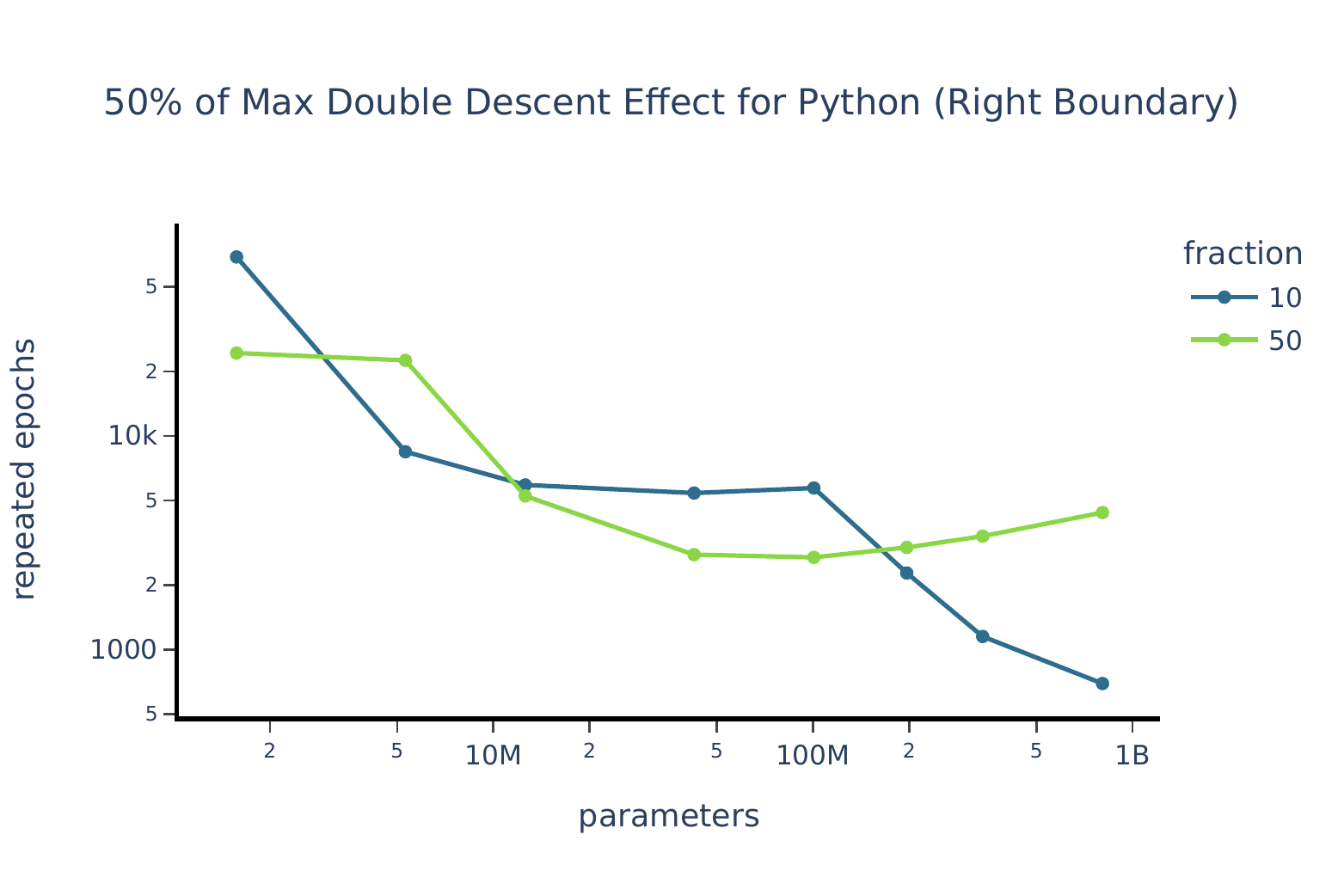}
    \caption{We estimate how many repeated epochs half would cause half of the maximum effect size (on a log scale) for our Python models  using linear interpolation on the left and right side of the double descent peak for each fraction of repeated data. We then averaged these two curves to make an overall estimate for the left and right boundaries of the poor performance region shown in Figure \ref{fig:py_double_descent}}
    \label{fig:py_region_fitting}
\end{figure}

For Python, the aggregate shown in Figure \ref{fig:py_double_descent} is quite a bit noisier. A lot of the noise is explained by only aggregating two scans rather than 5. But we see the individual scans for Python are also noiser as shown in Figure \ref{fig:py_region_fitting} 


\section{Appendix: Logit Attribution Analysis, 2 Layer Models}
\label{app:logit_attribution}
For attention only models we can directly attribute contributions of the attention heads to the logits. We attempted to use this technique to better understand how the induction heads were disrupted for 2 layer models. For instance, it could be they were firing more weakly, or it could be activity from other attention heads were interfering with their ability to copy.

\begin{figure}
    \centering
    \includegraphics[width=1.0\textwidth]{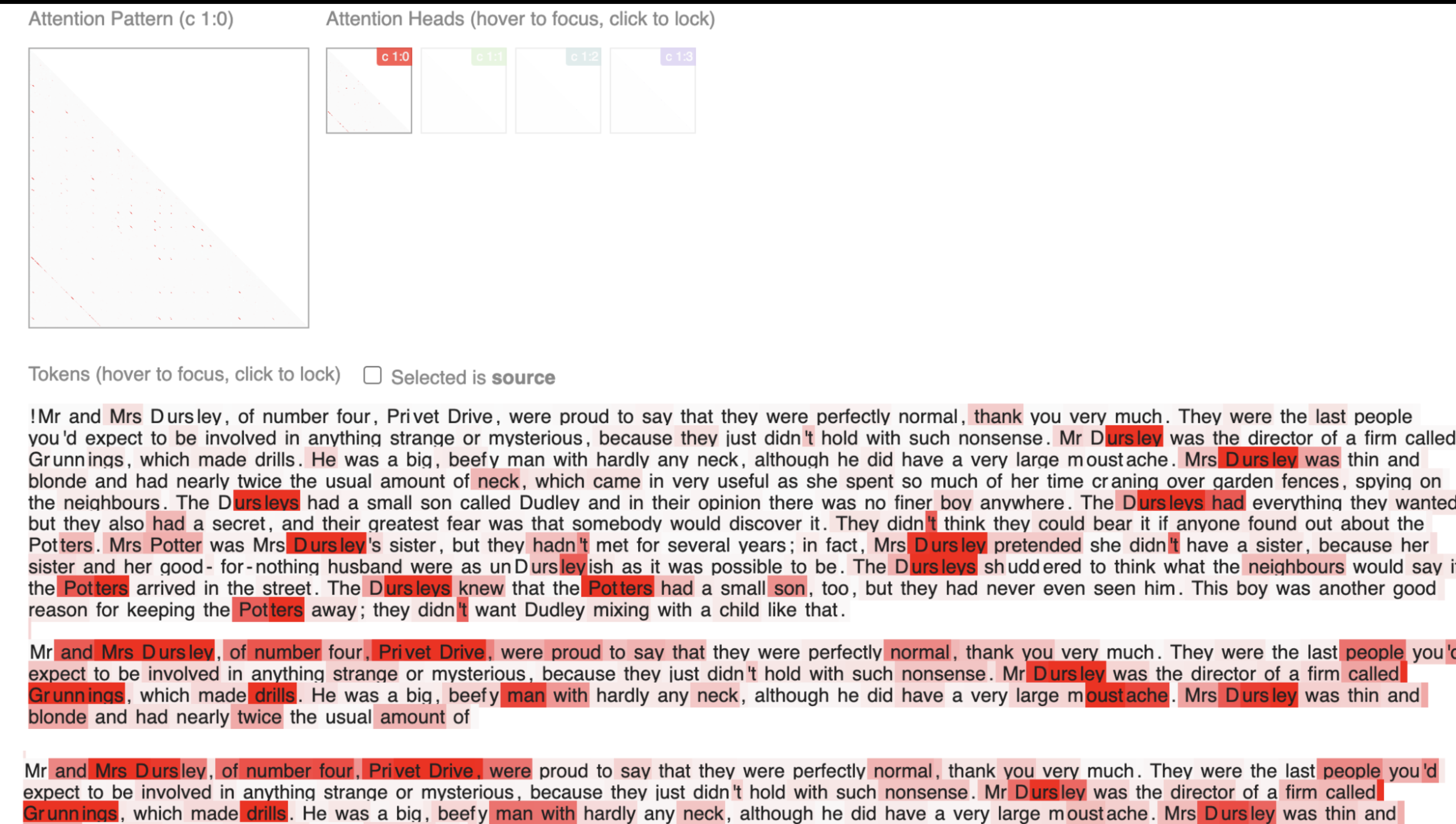}
    \caption{For attention only models we can directly attribute contributions of the attention heads to the logits as shown in \cite{elhage2021mathematical, olsson2022context}. Both models were evaluated on the first paragraph of Harry Potter copied twice. The induction head appeared to be head 0, shown in red for both models. The control model's logit attribution is shown for the first two paragraph, and the third paragraph shown is from the repeated model at the double descent peak for comparison.}
    \label{fig:logit_red}
\end{figure}

\begin{figure}
    \centering
    \includegraphics[width=1.0\textwidth]{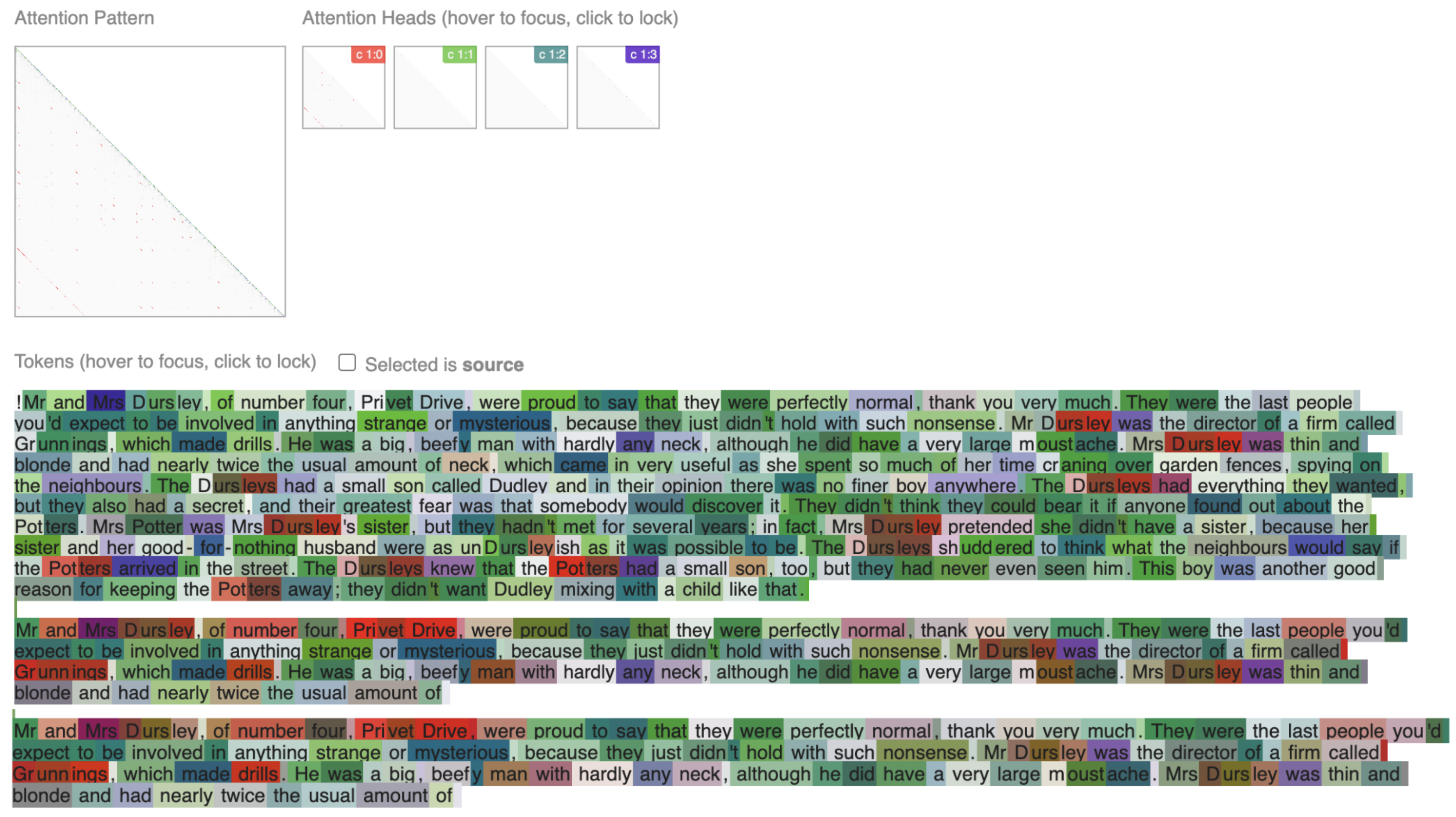}
    \caption{For attention only models we can directly attribute contributions of the attention heads to the logits as shown in \cite{elhage2021mathematical, olsson2022context}. Similar to Figure \ref{fig:logit_red} both models were evaluated on the first paragraph of Harry Potter copied twice, but here the contribution of all attention heads is shown. The other attention heads in the repeated data model appears more active (several of the reddish tokens in the second paragraph are brown in the third paragraph).}
    \label{fig:logit_all}
\end{figure}

Overall it feels like both effects happen weakly, and that it was easier to understand the disruption to induction heads through the per token losses shown in Figures \ref{fig:1l_attn_hp} and \ref{fig:2l_attn_hp}, than through logit attribution.

\begin{figure}
    \centering
    \includegraphics[width=.49\textwidth]{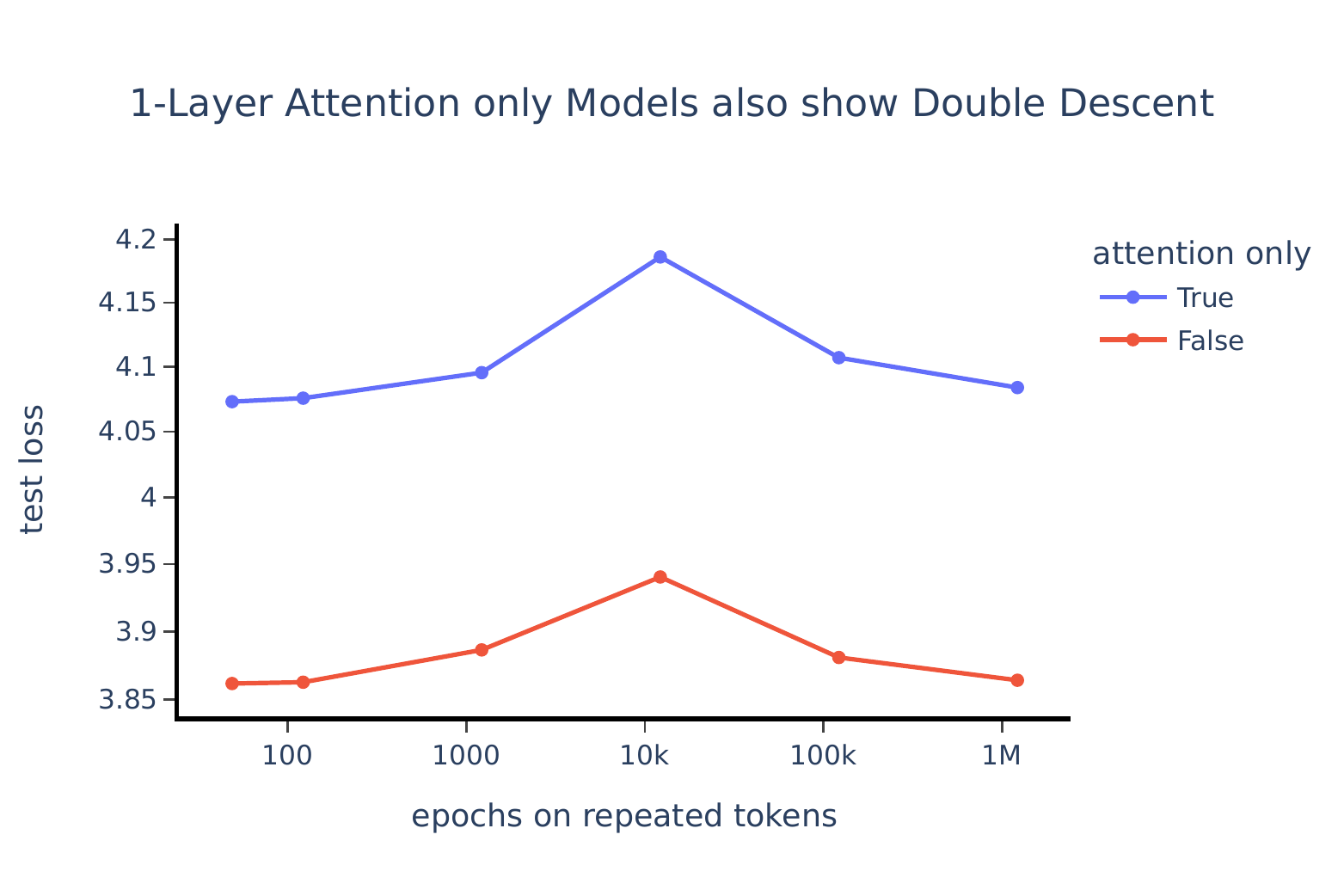}
    \caption{We still observe double descent on repeated data with 1 layer attention only models, so it is possible we'd observe double descent on repeated data for simpler model types.} 
    \label{fig:dd_1_layer}
\end{figure}

\newpage
\section{Appendix: Copying and Prefix Matching Score Fits}
\label{app:copy_prefix}

\begin{figure}[ht]
    \centering
    \includegraphics[width=0.49\textwidth]{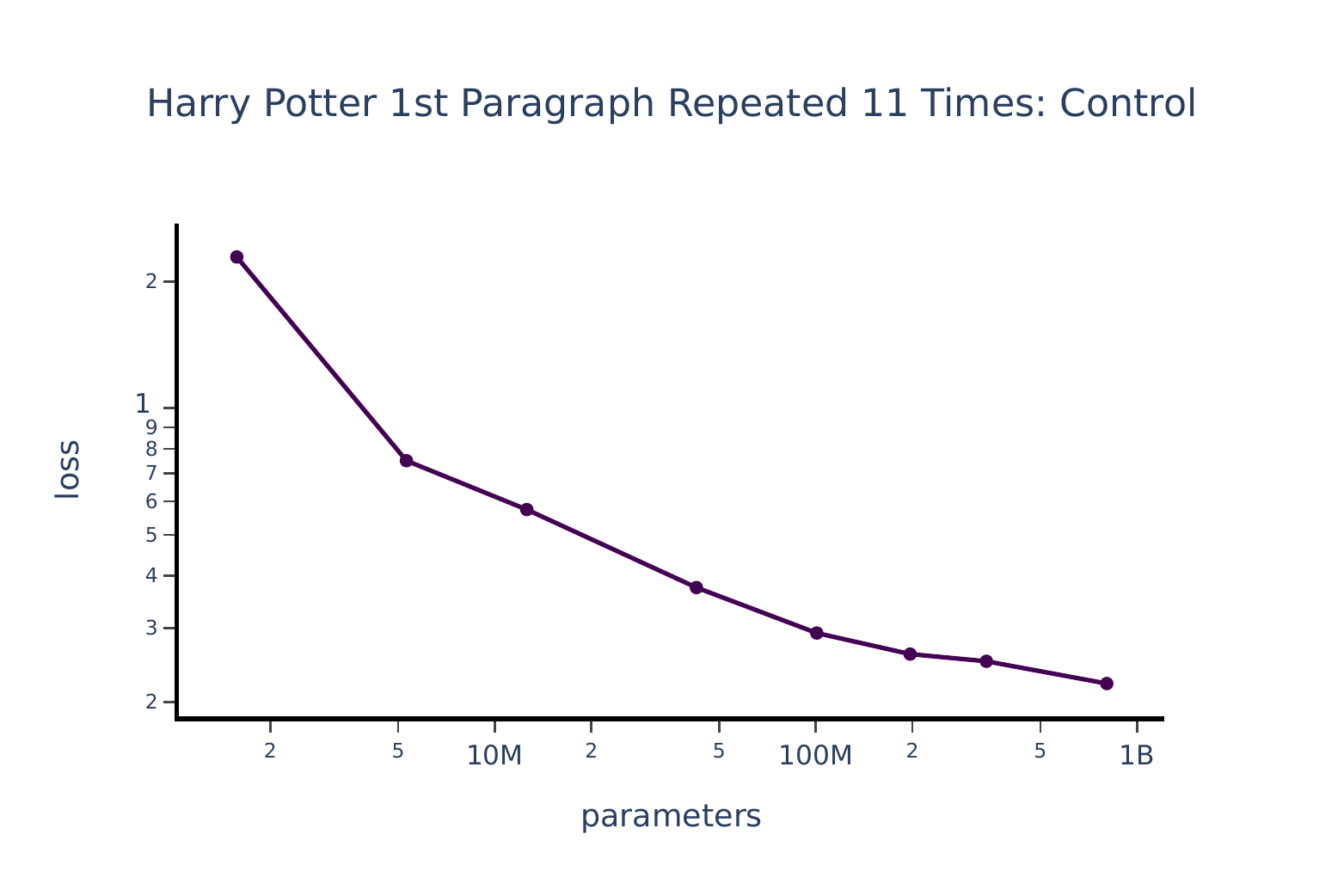}
    \includegraphics[width=0.49\textwidth]{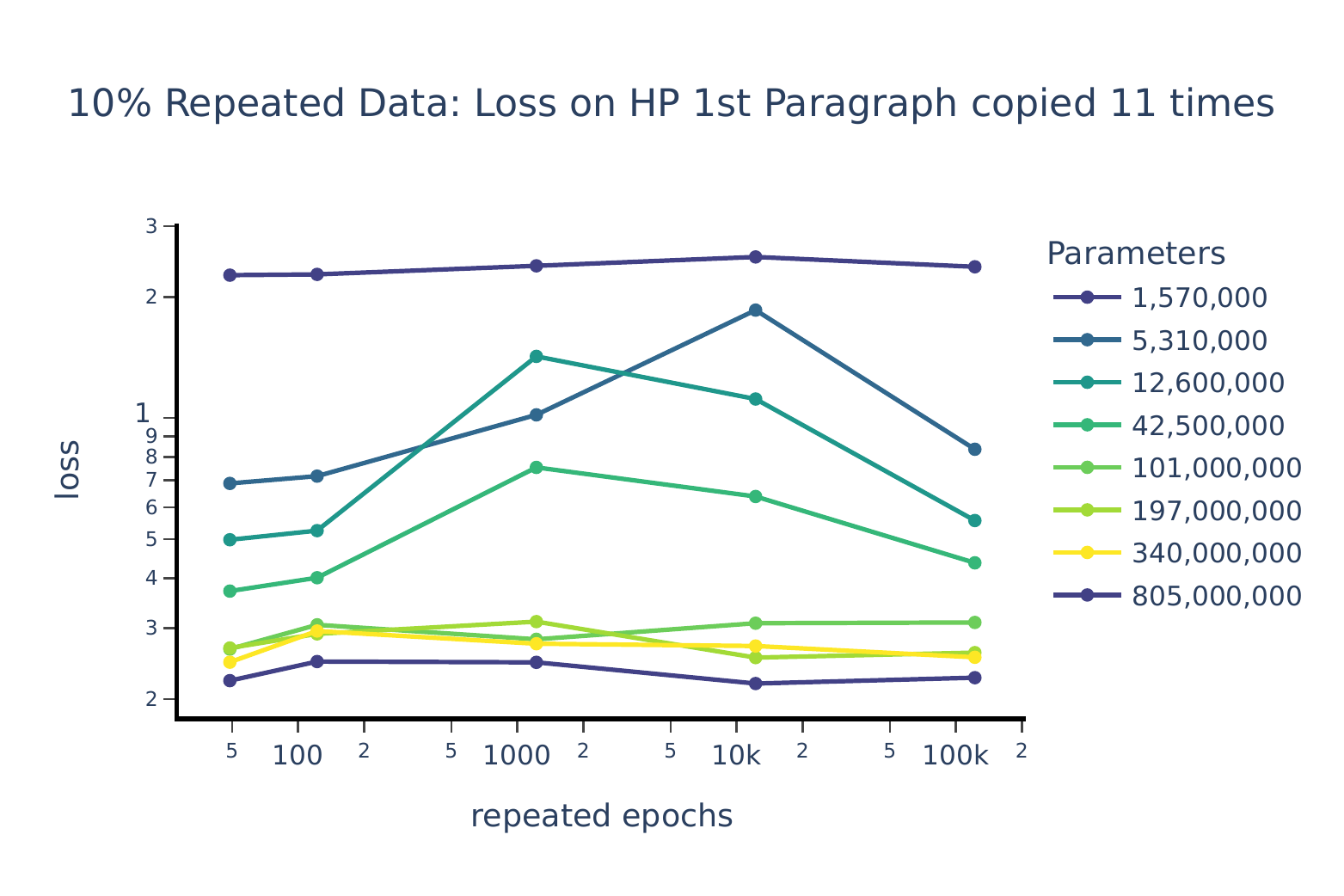}
    \caption{In order to do the model size interpolation used in Figure \ref{fig:hp_copying_1} we use the loss on Harry Potter's first paragraph copied 11 times for our control models (no repeated data). it is relatively well behaved, but it was not obvious how to extrapolate the curve. On the right, as a sanity check, we check to make sure we still see peaks moving left as model size increases that approximately line up with what was observed in \ref{fig:double_descent}} 
    \label{fig:appendix_c_hp}
\end{figure}

\begin{figure}[ht]
    \centering
    \includegraphics[width=0.49\textwidth]{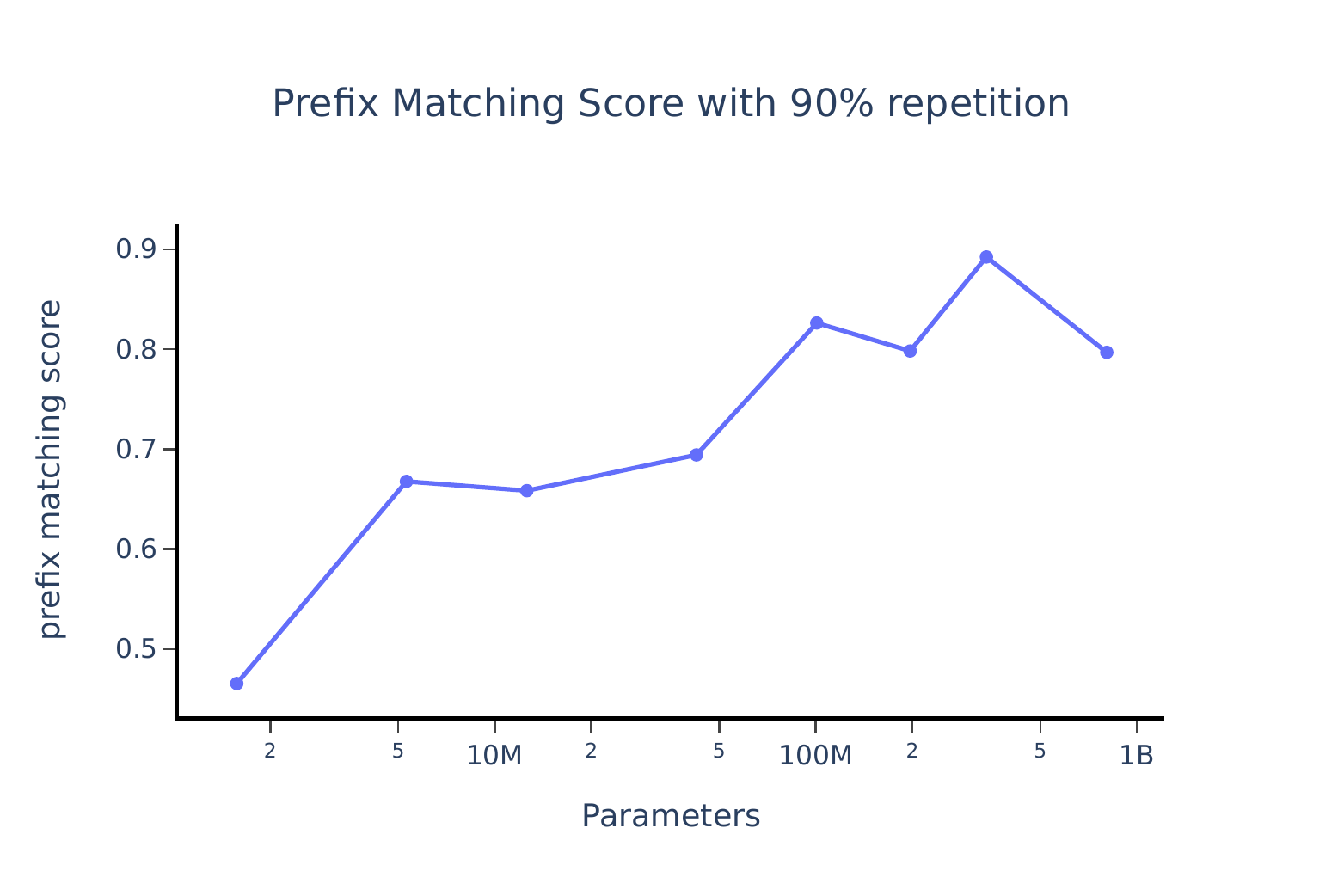}
    \includegraphics[width=0.49\textwidth]{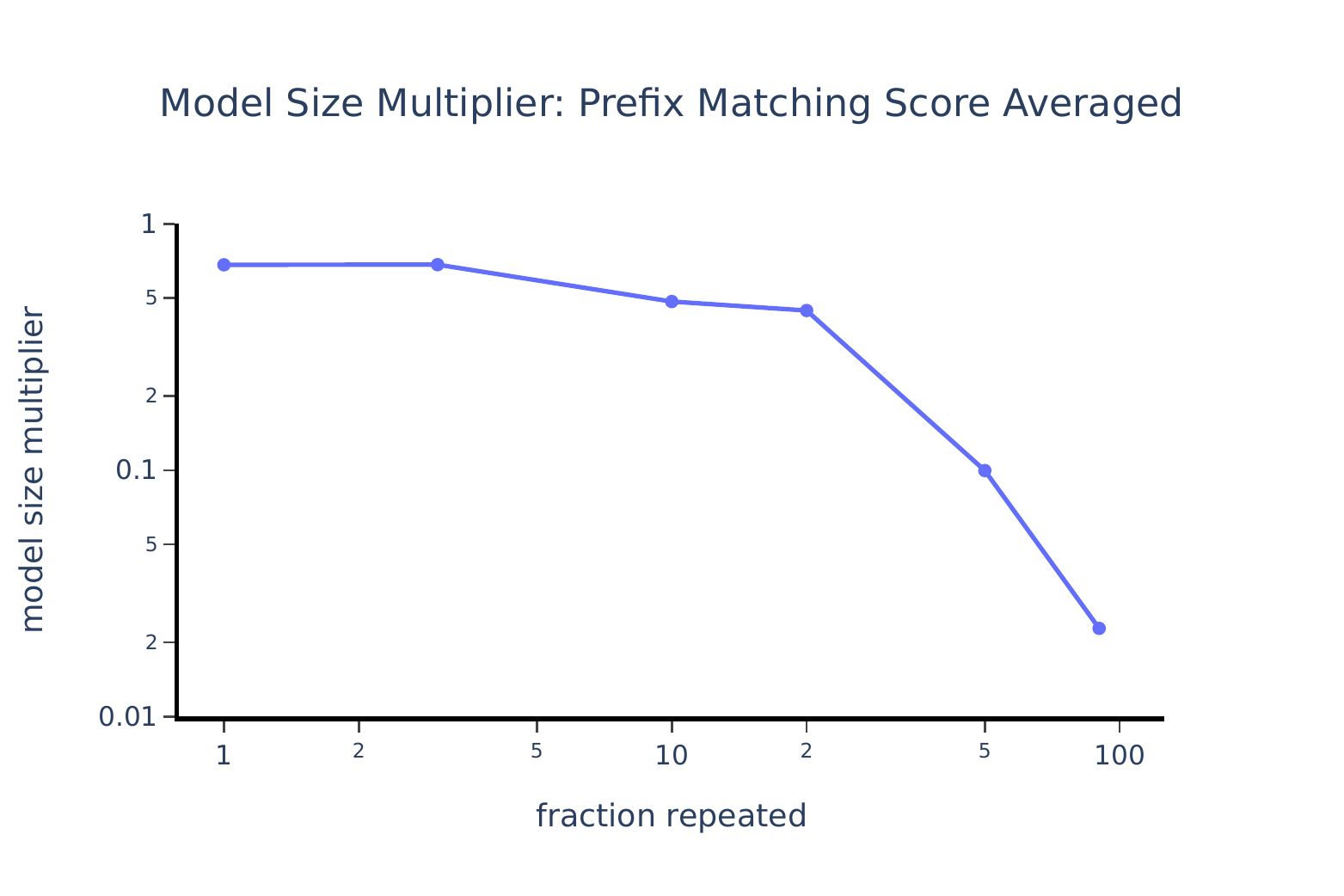}
    \caption{For the model size multiplier in Figure \ref{fig:prefix_matching} we a linear fit on the prefix matching score for the control models shown on the left. On the right, similar to \ref{fig:ratio_py_text} we show that if we take an average over model size (harmonic mean of multiplier), we get a relatively clean relationship.} 
    \label{fig:appendix_c_prefix}
\end{figure}

\newpage
\section{Appendix: Harry Potter Copying Evaluation with Fewer Characters}
\label{app:hp_fewer_characters}

\begin{figure}[ht]
    \centering
    \includegraphics[width=0.49\textwidth]{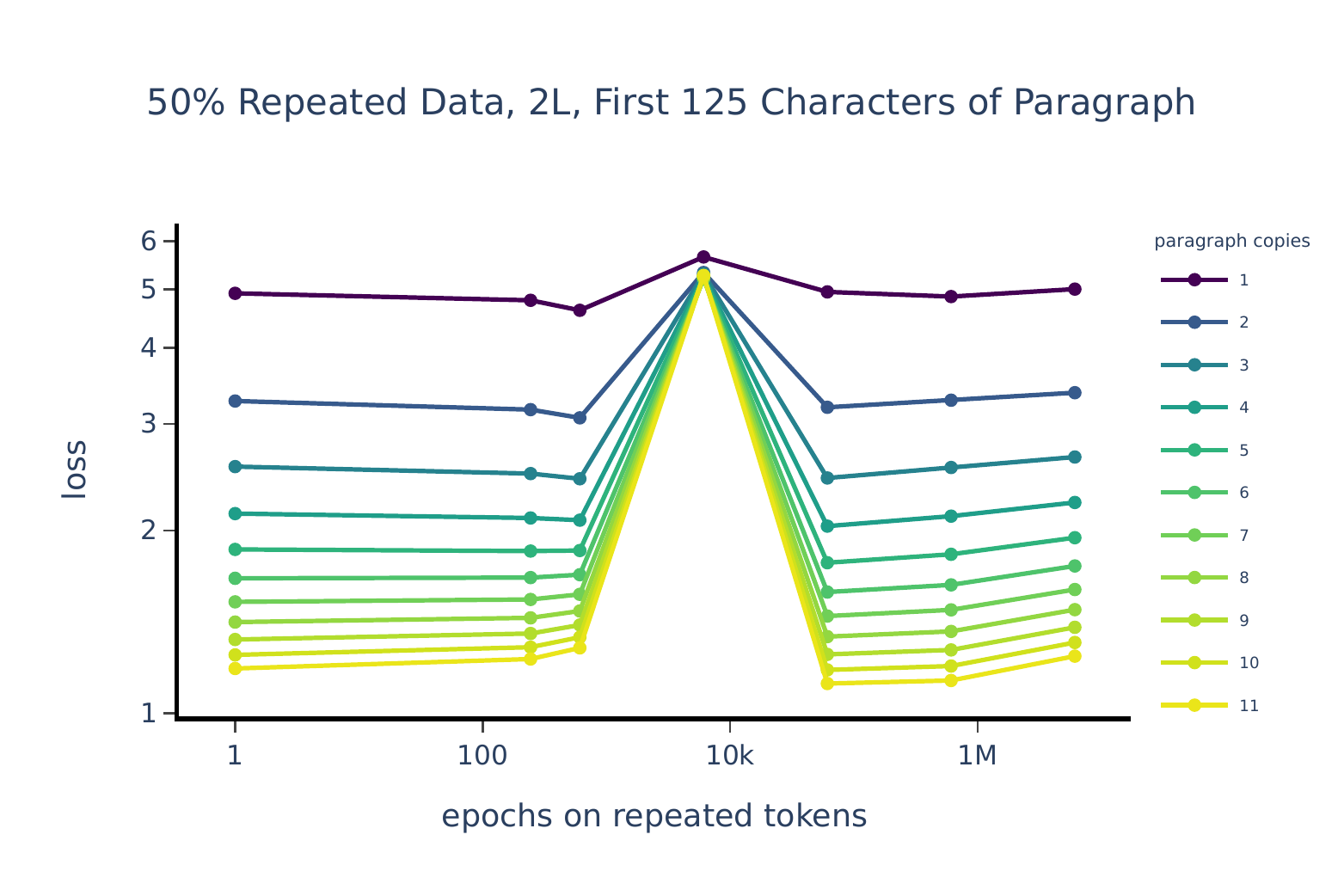}
    \caption{In order to make sure the copying eval was not merely evaluating in context learning, we tried a much shorter copied sequence (approximately 10x shorter, 125 characters instead of 1463). We still observe approximately no learning from repeated copying for the 2L model trained on 50\% repeated data at the double descent peak} 
    \label{fig:appendix_c_hp_broken_copying_fewer_characters}
\end{figure}


\clearpage

\bibliographystyle{apalike}
\bibliography{bibliography}

\end{document}